\newif\ifarxiv
\arxivtrue

\ifarxiv
  \documentclass{article}
  \usepackage[preprint]{colm2026_conference}
\else
  \documentclass[11pt]{article}
  \usepackage[review]{acl}
  \usepackage{times}
  \usepackage{latexsym}
\fi

\usepackage[T1]{fontenc}
\usepackage[utf8]{inputenc}
\usepackage{microtype}
\ifarxiv\else
  \usepackage{inconsolata}
\fi

\usepackage{amsmath}
\usepackage{amssymb}

\ifarxiv\else
  \usepackage{stfloats}        
  \usepackage{cuted}           
\fi
\usepackage{placeins}
\usepackage{float}
\usepackage{booktabs}

\usepackage{cleveref}

\usepackage{graphicx}
\graphicspath{{figures/}}
\ifarxiv
  \usepackage{capt-of}         
  \usepackage{etoolbox}
  \makeatletter
  \patchcmd{\l@section}{\addvspace{1.0em \@plus\p@}}{\addvspace{0.5em \@plus\p@}}{}{}
  \makeatother
\fi

\ifarxiv
  \newenvironment{widefigure}{\begin{figure}[H]\centering}{\end{figure}}
\else
  
\fi

\title{From Brewing into Resolution: Tracing \\ Internal Reasoning Trajectories in LLMs}

\ifarxiv
  \author{
    \textbf{Siyue Chen$^{*}$ \quad
    Yifu Guo$^{*}$ \quad
    Yuquan Lu \quad
    Zishan Xu \quad
    Jiaye Lin \quad
    Jianbo Lin} \\
    \textbf{Siyu Zhang \quad
    Cheng Yang \quad
    Junxin Li \quad
    Yujia Li \quad
    Yu Huo \quad
    Ruixuan Wang} \\[0.5em]
    {\fontsize{11}{14}\selectfont $^*$Core Contributors}
  }
  \githuburl{https://github.com/euyis1019/llm-brewing/tree/code_brewing_exp}
\else
  \author{First Author \\
    Affiliation / Address line 1 \\
    Affiliation / Address line 2 \\
    Affiliation / Address line 3 \\
    \texttt{email@domain} \\\And
    Second Author \\
    Affiliation / Address line 1 \\
    Affiliation / Address line 2 \\
    Affiliation / Address line 3 \\
    \texttt{email@domain} \\}
\fi

\begin{document}

\ifarxiv
  \newgeometry{left=1.2cm,right=1.2cm,top=1.2cm,bottom=1.2cm}
  \begin{abstract}
Although large language models (LLMs) have demonstrated promising performance across coding tasks, their behavior can still vary across semantically equivalent code snippets, where the same computation may produce different predictions.
However, the internal mechanisms underlying these differences remain unclear.
In this paper, we employ mechanistic analysis to investigate how internal answer representations evolve across model depth and why final predictions can conceal distinct trajectories.
We develop a hidden-state-based diagnostic framework that traces how representations of the correct answer evolve across model layers.
Our framework reveals three diagnostic notions that characterize how answer representations develop across model layers:
(1) \emph{Availability}: the correct answer is linearly recoverable from the hidden representation.
(2) \emph{Brewing}: the phase in which an available answer representation develops into a form the model can use.
(3) \emph{Readiness}: the correct answer is encoded in a form decodable by the model's own remaining computation.
Using this framework, we show that semantically equivalent code forms can diverge during the transition from Availability to Readiness, revealing internal differences that final accuracy alone cannot identify.
Applying the framework across coding tasks further reveals distinct internal bottlenecks across different reasoning primitives.
\end{abstract}
  \maketitle                       
  \restoregeometry
  \begingroup
    \small                         
    \tableofcontents
  \endgroup
  \newpage
\else
  \maketitle
  \begin{abstract}
Although large language models (LLMs) have demonstrated promising performance across coding tasks, their behavior can still vary across semantically equivalent code snippets, where the same computation may produce different predictions.
However, the internal mechanisms underlying these differences remain unclear.
In this paper, we employ mechanistic analysis to investigate how internal answer representations evolve across model depth and why final predictions can conceal distinct trajectories.
We develop a hidden-state-based diagnostic framework that traces how representations of the correct answer evolve across model layers.
Our framework reveals three diagnostic notions that characterize how answer representations develop across model layers:
(1) \emph{Availability}: the correct answer is linearly recoverable from the hidden representation.
(2) \emph{Brewing}: the phase in which an available answer representation develops into a form the model can use.
(3) \emph{Readiness}: the correct answer is encoded in a form decodable by the model's own remaining computation.
Using this framework, we show that semantically equivalent code forms can diverge during the transition from Availability to Readiness, revealing internal differences that final accuracy alone cannot identify.
Applying the framework across coding tasks further reveals distinct internal bottlenecks across different reasoning primitives.
\end{abstract}

\fi

\section{Introduction}

Large language models have become increasingly capable in code reasoning, yet semantically equivalent programs can still produce different predictions \citep{rabin2021generalizability}. Recent work on semantics-preserving code mutations reports that loop unrolling causes the largest accuracy drop among the transformations studied, even for otherwise strong models \citep{orvalho2025large}. Accuracy, however, only tells us whether the final answer is correct; it does not reveal where internal trajectories separate or which stage of computation causes the difference.

A common way to examine internal states is to apply a layer-wise readout that tests whether the answer can be recovered from the hidden state at each layer. Linear probing fits an external classifier for this purpose \citep{hewitt2019structural, belinkov2022probing}, while vocabulary lenses instead project the hidden state through the model's unembedding, either directly or through a learned affine translator \citep{nostalgebraist2020logitlens, belrose2023tunedlens}. Both families share a common characteristic: they substitute a readout of their own for the computation the model would otherwise perform above that layer. Even when such a readout recovers the answer, this does not establish that the model can decode it from the same state through its remaining computation \citep{ravichander2021probing, belinkov2022probing}. Layer-wise readouts therefore characterize whether answer information is recoverable from a hidden state, while leaving open whether the model itself can use that state. This distinction matters because a hidden state may contain answer information without yet providing a representation that the remaining layers can directly transform into the final prediction. Conversely, a representation that already supports decoding may still be altered by later computation before the output is produced. Therefore, observing whether an answer can be extracted is insufficient to determine whether the model has completed the corresponding computation.

To capture this distinction, we separate two aspects of how a hidden state relates to the correct answer. Availability asks whether the answer can be recovered from the hidden state by an external readout, while Readiness asks whether the model's own remaining computation can decode the answer after the original code context is removed. Together, they provide a way to locate where internal trajectories diverge across model depth.

To measure both aspects at every layer, we introduce a paired diagnostic framework. We measure Availability with linear probing, using a simple external classifier to recover the answer from each hidden state. We measure Readiness with Context-Stripped Decoding, which builds on the activation patching approach of \citet{ghandeharioun2024patchscopes}. CSD transfers the same hidden state into a target prompt from which the original code context has been removed, then continues the forward pass through the model's remaining layers so that Readiness is judged by the model's own computation rather than an external readout. The resulting prediction tests whether those layers can decode the correct answer from the transferred state without access to the original code. Applying the two diagnostics across model depth produces paired trajectories for each sample.

The paired trajectories reveal Brewing, a recurring interval in which the answer has become available but has not yet reached Readiness. This interval lets us distinguish trajectories that separate after Availability, including those that fail to reach Readiness and those that fail only after Readiness has been reached. Across task families and models, the paired diagnostics give a common coordinate system for analyzing how different code structures affect internal computation.

\begin{figure}[t]
\centering
\includegraphics[width=\columnwidth]{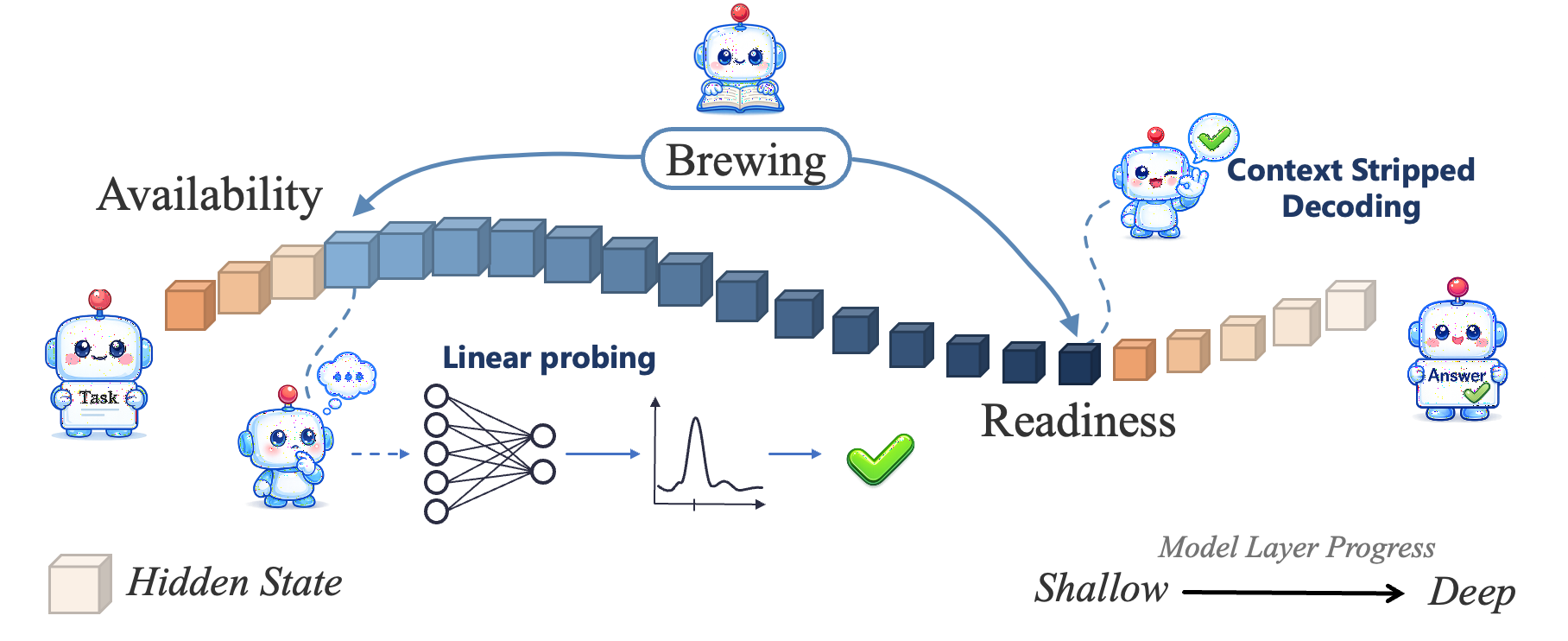}
\caption{
Illustration of answer formation across model layers. The blue region
highlights the Brewing phase between Availability and Readiness.
}
\label{fig:brewing_overview}
\end{figure}

We apply this coordinate system to the loop unrolling gap that motivates our study. Using strictly matched Loop and Unrolled programs, the framework localizes their divergence not to the onset of Availability, which the two forms reach at comparable points, but to the subsequent transition toward Readiness. On the anchor model, we further trace this difference to late MLP consolidation in the dual-variable condition, showing how the framework connects a behavioral difference to a specific internal computation.

Our main contributions are as follows.

\begin{itemize}
\item We distinguish Availability and Readiness as two aspects of how a hidden state relates to the correct answer, and identify Brewing as the recurring phase in which the answer has reached Availability but has not yet reached Readiness.

\item We introduce a paired diagnostic framework that measures both aspects across model depth, combining linear probing for Availability with Context-Stripped Decoding for Readiness to trace answer trajectories inside the model.

\item We use the framework to locate where code structures affect internal computation. Applied to strictly matched Loop and Unrolled programs, it localizes their divergence to the transition toward Readiness and identifies a late MLP mechanism in the anchor model under the dual-variable condition.
\end{itemize}

\section{Related Work}

\subsection{Analysis of Hidden Representations}

Prior research uses intermediate hidden states to study code correctness and
whether a generation will ultimately succeed
\citep{bui2025openia, afzal2025knowing, ribeiro2026internal,
dicicco2026decodable}. Related work
also localizes code-relevant representations across layers, token positions,
and neurons \citep{vu2026autoprobe, yin2026neuronguided}. Code-specific probing
has examined the extent to which pretrained code models capture syntactic
structure, identifier-related information, and semantic equivalence
\citep{troshin2022probing}. Linear probing, Logit Lens, and Tuned Lens provide
common ways to read information from hidden states at different depths
\citep{hewitt2019structural, belinkov2022probing, nostalgebraist2020logitlens,
belrose2023tunedlens}.

Probe performance can partly reflect the probe's capacity to learn the task
rather than information encoded by the representation, motivating control-task
evaluations \citep{hewitt2019designing}.
These methods reveal when answer information becomes recoverable from hidden
states, but recovery through an external readout does not specify whether the
model's own remaining computation can decode the answer from the same state
\citep{ravichander2021probing, belinkov2022probing}.

\subsection{Decoding Intermediate Representations}

A related line of work analyzes how hidden states develop across depth rather
than reading them at a single layer. Geometric analyses track the intrinsic
dimension and manifold structure of representations along the forward pass and
identify distinct phases in how information is organized
\citep{valeriani2023geometry, cheng2025abstraction, ma2026manifolds}.
Complementary analyses characterize transformer feed-forward layers as
key--value memories and trace how their updates promote concepts into vocabulary
space as predictions develop across depth
\citep{geva2021kv, geva2022transformer}. A second strand feeds an intermediate
state back into a language model, either the same model or a different one, and
reads the resulting generation
\citep{ghandeharioun2024patchscopes, chen2024selfie}.

Geometric analyses describe how representations are organized without reference
to a specific answer, while generation-based decoding reports what a state
encodes through a single readout at each layer. In both cases the depth at
which an answer appears is relative to the readout that was chosen. We instead
hold the hidden state fixed and vary only the readout condition: an external
linear map for Availability, and the model's own remaining computation under a
stripped context for Readiness. Because CSD retains the question suffix and
removes only the code, its prediction lies in the same target space as the
probe at the same layer, so the two are directly comparable. The quantity we
analyze is the interval between them, which no single readout reports however
accurate it is, and which the paired diagnostic in \Cref{sec:framework} makes
measurable.

\subsection{Mechanistic Analysis across Layers and Components}

A third line of work moves from layer-wise states to individual components.
Layer-wise readouts can locate the depth at which two trajectories separate,
but cannot identify the operations that produce the separation. Attribution
graphs, causal tracing, and related circuit-discovery methods
\citep{meng2022locating, ameisen2025circuit,
wang2023ioi, conmy2023automated} typically begin
from an observed output and trace it to the features, attention heads, MLPs,
and connections that support it. Causal-abstraction work provides a framework
for testing whether such component-level accounts remain faithful under
intervention \citep{geiger2025causal}.
Our analysis shifts the attribution target from a final prediction to the
divergence between matched internal trajectories. Locating this separation
across depth requires comparable readouts of the same answer state, which the
paired diagnostic in \Cref{sec:framework} provides.

\section{Paired Diagnostic Framework}
\label{sec:framework}

We introduce a paired diagnostic framework that traces how an answer state
changes across model depth. The framework separates two aspects of a hidden state, whether
the answer can be recovered from the state and whether the model itself can use
the state through its remaining computation.

\subsection{Controlled Reasoning Setup}
\label{sec:setup}

We study matched pairs of short, deterministic Python programs. The two programs
in each pair receive the same inputs and produce the same answer, while one
uses a \texttt{for} loop and the other expresses the same computation as
explicit sequential statements. Across pairs, we vary the loop body, iteration
count, and initial values, randomize identifiers, and execute every program to
verify its answer.

Answers are restricted to a single-digit token in
$\mathcal{D}=\{0,\ldots,9\}$. This provides a shared target space across models
and removes output length variation from layer-wise comparison. Complementary
results on the execution benchmark CRUXEval-O \citep{gu2024cruxeval} are reported in
\Cref{app:cruxeval}.

We use Qwen2.5-Coder-7B \citep{hui2024qwen2} as the anchor model for the matched comparison and all
mechanistic analyses. We additionally apply the paired diagnostic across five
other task families and fifteen additional models spanning the Qwen, Llama, and
DeepSeek families. Each task family contains 27 configurations and 4,050
instances, yielding 24,300 programs across the six families. Full experimental
settings appear in \Cref{app:models,app:bench}.

\subsection{Measuring Availability and Readiness}
\label{sec:diagnostics}

For each program, we write the source prompt as $S=[C;Q]$, where $C$ is the
code snippet and $Q$ is the question suffix. The correct answer is denoted by
$t^{*}\in\mathcal{D}$. For a decoder-only transformer with $L$ layers, let
$h^{\ell}\in\mathbb{R}^{d}$ denote the hidden state at the final prompt
position after layer $\ell$.

We measure answer \emph{Availability} with layer-wise Probing. At each layer, we
fit an independent multinomial linear classifier

\begin{equation}
    \Phi^{\ell}_{P}(h^{\ell})
    =
    \operatorname{softmax}
    \left(
        W_{\ell}h^{\ell}+b_{\ell}
    \right).
    \label{eq:probe}
\end{equation}

Here, $W_{\ell} \in \mathbb{R}^{|\mathcal{T}| \times d}$ and
$b_{\ell} \in \mathbb{R}^{|\mathcal{T}|}$ are the weight and bias of the classifier at
layer $\ell$. The softmax returns a distribution over the shared target space
defined below.

A layer is Probing correct when the classifier assigns its highest probability
to $t^{*}$. This indicates that the answer is encoded in a linearly recoverable
form in the hidden state.

Both diagnostics use the shared target space
$\mathcal{T}=\mathcal{D}\sqcup\{\bar d\}$. The ten digit classes represent
candidate answers, while the residual class $\bar d$ provides an alternative
when no single-digit answer is linearly recoverable. It prevents early hidden
states from being forced into an arbitrary digit prediction. The construction
of $\bar d$ is described in \Cref{app:probe}. Using the same target space
keeps the candidate answers fixed, so disagreement between Probing and CSD can
be attributed to their different readout conditions.

Building on Patchscopes \citep{ghandeharioun2024patchscopes}, we measure answer
\emph{Readiness} with Context-Stripped Decoding (CSD). CSD
tests the same hidden state through the model's own remaining computation after
removing the original code context. We construct a target run containing only
$Q$, replace the hidden state at the final position after layer $\ell$ with
the source state $h^{\ell}$, and continue the forward pass through the remaining
layers. The source and target runs remain separate at every layer. The source
run supplies $h^\ell$, while the target run supplies the stripped context used
by the remaining computation. Only the final-position state is replaced, which
keeps the CSD condition comparable across layers.

The stripped target prompt can induce a digit preference even without hidden
state transfer. We therefore run the target prompt without patching to obtain
baseline logits $z_{\mathrm{base}}$ and subtract them from the patched logits:

\begin{equation}
    \Phi^{\ell}_{C}(h^{\ell})
    =
    \operatorname{softmax}_{\mathcal{T}}
    \left(
        z^{\ell}_{\mathrm{patch}}
        -
        z_{\mathrm{base}}
    \right).
    \label{eq:csd}
\end{equation}

Here, $z^\ell_{\mathrm{patch}}$ denotes the logits produced by the patched
target run, while $z_{\mathrm{base}}$ denotes the logits produced by the same
target prompt without hidden-state transfer. The subtraction removes the preference introduced by the target prompt and
isolates the contribution of the transferred hidden state. A layer is CSD
correct when $\Phi^{\ell}_{C}(h^{\ell})$ assigns its highest probability to
$t^{*}$.

The two diagnostics impose different readout conditions on the same hidden
state. Probing applies an external linear map and tests whether the answer is
linearly recoverable from the hidden state. CSD relies on the model's
remaining computation under the stripped context and tests whether those layers
can decode the answer from the transferred state.

When Probing becomes correct before CSD, the answer has reached Availability
but has not yet reached Readiness. When CSD becomes correct at a layer where
Probing is already correct, both readouts are correct for the same hidden
state. These layer-wise predictions over $\mathcal{T}$ are distinct from the four trajectory
outcomes in \Cref{sec:outcomes}. The predictions describe one hidden state,
while the outcomes summarize whether joint correctness is reached and preserved
across the full forward pass.

We observe that Probing typically becomes correct several layers before CSD
across the six task families. Probing first becomes correct no later than CSD
in 74.2\% of samples. To measure this pattern across model depth, we introduce
two landmarks below. Additional implementation details
appear in \Cref{app:diagnostic}, and validity checks for CSD appear in
\Cref{app:csd-validity}.

\subsection{Landmarks}
\label{sec:landmarks}

To characterize the transition observed above, we use these trajectories to
define two landmarks that locate when the correct answer first reaches
Availability and Readiness.

Let $\hat t^{\ell}_{P}$ and $\hat t^{\ell}_{C}$ denote the argmax predictions
of Probing and CSD at layer $\ell$. The First Probing Correct Layer and the
First Joint Correct Layer are defined as

\begin{align}
    \operatorname{FPCL}(S)
    &=
    \min
    \left\{
        \ell
        \,\middle|\,
        \hat t^{\ell}_{P}=t^{*}
    \right\},
    \label{eq:fpcl}
    \\
    \operatorname{FJC}(S)
    &=
    \min
    \left\{
        \ell
        \,\middle|\,
        \hat t^{\ell}_{P}=t^{*}
        \land
        \hat t^{\ell}_{C}=t^{*}
    \right\}.
    \label{eq:fjc}
\end{align}

FPCL marks the first layer where the answer becomes linearly recoverable and
therefore provides the operational landmark for Availability. FJC marks the
first layer where Probing and CSD are jointly correct and provides the
corresponding landmark for Readiness. This correspondence follows from how
the two diagnostics read the same hidden state. Probing applies an external
linear map and tests whether the answer is linearly recoverable, while CSD
continues the model's own remaining computation and tests whether those layers
can decode the answer after context removal.

Because the definition of FJC includes Probing correctness, every defined FJC
occurs at or after FPCL. When both landmarks exist, they therefore provide a
well-defined interval between the first attainment of Availability and the
first attainment of Readiness. Section~\ref{sec:brewing} uses this interval to
define Brewing.

Both landmarks record first attainment. FJC identifies the earliest jointly
correct state without requiring that this state remain correct in subsequent
layers. Whether the answer is preserved after FJC is evaluated separately
through the final model prediction. This separation allows later analysis to
distinguish trajectories that preserve a jointly correct state from those in
which subsequent computation moves away from it. The main findings remain stable when
landmark attainment requires correctness for two or three consecutive layers
(\Cref{app:landmark-persistence}).

If no qualifying layer exists, the corresponding landmark is recorded as
$\varnothing$. Samples without FPCL are reported separately. In these samples, the correct answer never becomes
linearly recoverable at any evaluated layer, so Availability is not reached
under the Probing diagnostic. Samples with FPCL but without FJC reach
Availability, but Probing and CSD never become jointly correct within the
available model depth. The transition toward Readiness therefore begins but
does not complete. Section~\ref{sec:outcomes} further separates these
FJC-undefined trajectories according to their late-layer CSD behavior.

\section{Brewing and Resolution}
\label{sec:brewing}

\subsection{The Brewing Phase}
\label{subsec:brewing_phase}

The two diagnostics reveal a consistent separation between when an answer can
be extracted and when it can be used by the model itself. When both landmarks
exist, we define the interval
$[\operatorname{FPCL}(S),\operatorname{FJC}(S))$
as \emph{Brewing}. During this phase, the answer is available from the hidden
state but has not yet reached a form that the model can decode through its own
remaining layers. We define the Brewing duration as

\begin{equation}
    \Delta_{\mathrm{brew}}(S)
    =
    \operatorname{FJC}(S)
    -
    \operatorname{FPCL}(S).
\end{equation}

When FPCL exists but FJC is undefined, the transition begins but does not
complete within the available depth. When FPCL is undefined, the sample is
assigned to \textsc{No Brewing}.

Brewing appears across all six task families and sixteen evaluated models. In
the anchor model, the average duration is 10.7 layers, corresponding to
approximately 38\% of model depth. Across models, normalized Brewing duration
ranges from 24\% to 42\%. These results show that the gap between answer
Availability and Readiness persists across different tasks and models.
Detailed cross-task and cross-model results appear in
\Cref{app:crossmodel}.

\subsection{Four Resolution Outcomes}
\label{sec:outcomes}

\begin{figure*}[t]
  \centering
  \includegraphics[width=0.92\textwidth]{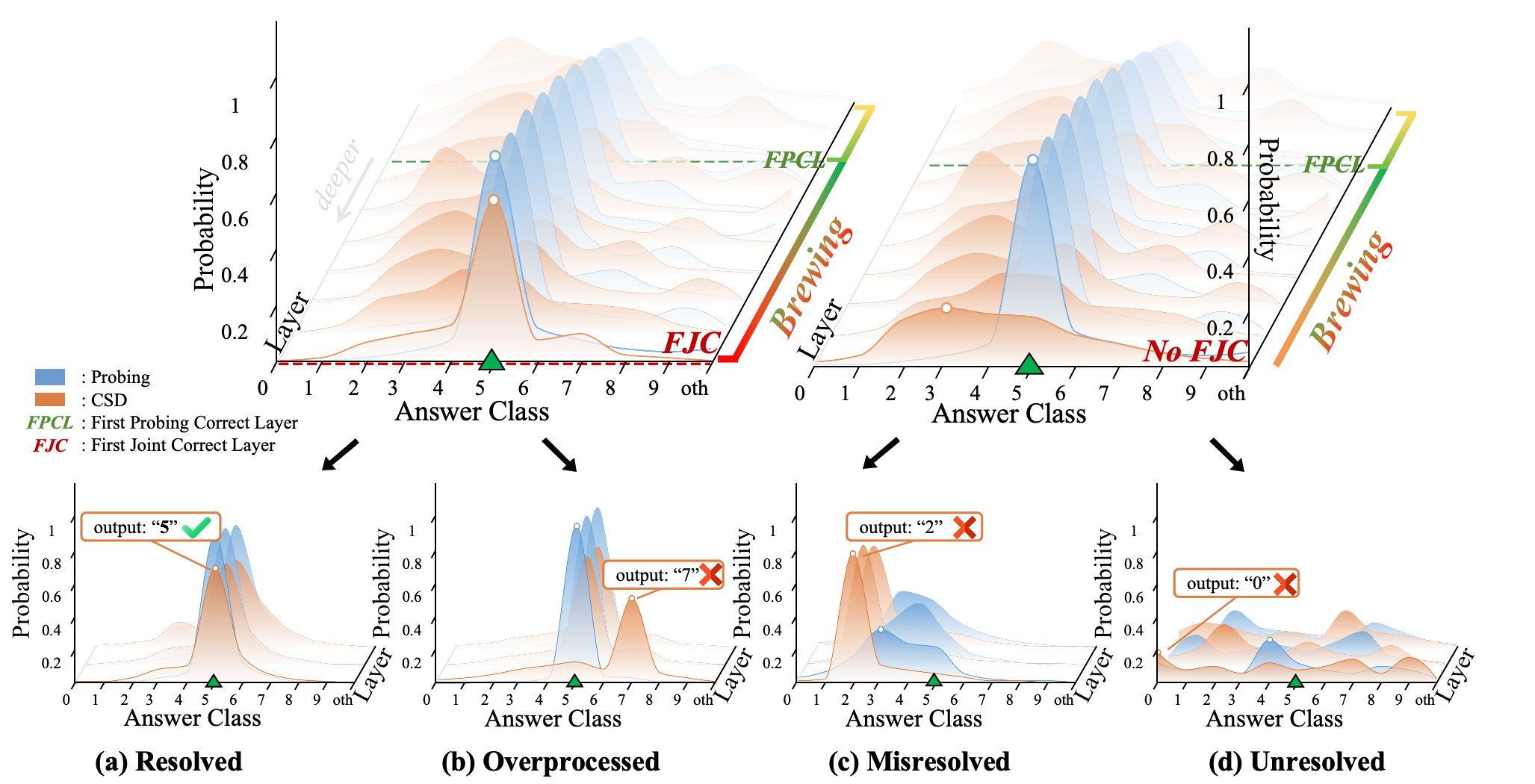}
  \caption{
  Paired diagnostic trajectories and resolution outcomes.
  FPCL marks the first correct Probing layer, and FJC marks the first layer of
  joint Probing and CSD correctness. Trajectories with or without FJC produce the
  four outcomes: Resolved, Overprocessed, Misresolved, and Unresolved.
  }
  \label{fig:outcomes}
\end{figure*}

Let $\hat t_{\mathrm{out}}$ denote the final model prediction. We classify
each trajectory according to whether joint correctness is reached and whether
it is preserved.

For samples with defined FJC, a sample is \textsc{Resolved} when
$\hat t_{\mathrm{out}} = t^{*}$, and \textsc{Overprocessed} otherwise.

For samples without FJC, we compute the average CSD probability over the final
quarter of layers,
$\overline{\Phi}^{W}_{C}(t) = \frac{1}{|W|} \sum_{\ell \in W} \Phi^{\ell}_{C}(h^{\ell})[t]$,
where $W = \{\ell \mid \ell \geq \lfloor 3L/4 \rfloor\}$.
A sample is classified as \textsc{Misresolved} when
$\max_{t \in \mathcal{D}} \overline{\Phi}^{W}_{C}(t) \geq 0.5$,
and as \textsc{Unresolved} otherwise.

Resolved and Overprocessed samples both reach FJC. They differ in whether the
jointly correct state is preserved in the final prediction.

Misresolved and Unresolved samples do not reach FJC. Misresolved samples form a stable late CSD preference for an incorrect digit, whereas Unresolved samples do not develop a comparable late-layer candidate.
FJC absence therefore places both outcomes before Readiness. Their late-layer
CSD behavior then separates concentration on one incorrect digit from a
trajectory that does not concentrate on any candidate. The threshold and tail-window choices are analyzed in \Cref{app:thresholds}.

Together, these outcomes separate trajectories that fail before Readiness from
trajectories that diverge after Readiness has been reached. They provide a
trajectory-level view of how an answer state can fail to reach, preserve, or
move away from a correct resolution. These distinctions motivate the functional
validation in \Cref{sec:validation}, where we test whether these answer states
respond differently under intervention.

Although the taxonomy uses the executed answer, \Cref{app:gt-free} additionally
evaluates ground-truth-free resolution detection from late-layer signals.

\subsection{Functional Validation}
\label{sec:validation}

We next test whether the states identified by the paired diagnostics have
predictable consequences under intervention. Because FPCL, FJC, and the four
outcomes are defined from Probing and CSD trajectories, their diagnostic
definitions alone do not establish how the corresponding hidden states affect
subsequent computation. We therefore examine the transition at FJC, the
preservation failure in Overprocessed trajectories, and the incomplete
transition in Unresolved trajectories.

\paragraph{Transfer at FJC.}
We test whether FJC coincides with a change in the usability of the
answer representation. We transfer hidden states from layers around FJC into a
neutral target prompt and allow the remaining layers to generate the next token
over the full vocabulary. Across the six task families, the correct answer
transfer rate rises from 3--18\% at layers before FJC to 20--44\% at FJC, with
an average increase of 21.2 percentage points. The transfer rate remains
elevated at subsequent layers. This rise is measured under a separate target prompt with full-vocabulary
decoding, so it does not follow from the FJC definition itself. States at and
after FJC are more readily decoded under this separate condition, and FJC
therefore aligns with a transition in how effectively the remaining computation
can use the transferred hidden state.

\paragraph{Layer skipping for Overprocessed.}
We examine the preservation failure captured by Overprocessed
trajectories. These samples reach FJC but eventually produce an incorrect final
answer, indicating that a jointly correct state was formed but not preserved.
To test whether the earlier representation can still guide the output, we
reintroduce the FJC hidden state at a downstream layer of the same forward pass
through alpha-blend injection. At the nearest downstream target layer, this
intervention restores the correct answer for 47.8\% of Overprocessed samples on
average across tasks. Resolved control trajectories retain higher accuracy
under the same intervention, which argues against a general correction effect
from alpha-blend injection. A usable answer representation had therefore formed
before the final error in these samples. Restoring that representation
downstream can counteract part of the later computation that moves the
trajectory away from the correct answer. This result supports interpreting
Overprocessed as a failure of preservation after Readiness. The remaining
failures show that reintroducing one earlier state does not fully reverse later
processing in every trajectory.

\paragraph{FPCL re-injection for Unresolved.}
We test whether Unresolved trajectories retain information about the
correct answer despite never reaching FJC. These samples have a defined FPCL,
so the correct answer becomes linearly recoverable, but the original forward
pass does not convert that representation into joint correctness. We inject the
FPCL hidden state into the final few layers and recover the correct answer for
22--38\% of Unresolved samples across tasks. Under the same intervention,
Resolved control samples retain 84--100\% accuracy, showing that the
intervention does not generally disrupt trajectories that have already
resolved. For a subset of Unresolved trajectories, the FPCL hidden state
therefore contains information that can still support the correct prediction
when introduced closer to the output, although the unmodified computation does
not complete the transition to Readiness. The partial rescue also limits this
interpretation, since many Unresolved samples remain incorrect after
re-injection and may require additional transformations beyond what the FPCL
hidden state provides.

Taken together, the three interventions show that the diagnostic categories
correspond to states with different responses to intervention. Transfer around
FJC supports its interpretation as a landmark in the transition to Readiness.
Reintroducing the FJC hidden state links Overprocessed trajectories to a
preservation failure after Readiness has been reached. Re-injecting the FPCL
hidden state shows that some Unresolved trajectories contain available answer
information without completing the transition toward Readiness. These results
show that the outcome taxonomy distinguishes internal trajectories that cannot
be separated from their final answers alone. Full intervention settings, layer
sweeps, injection modes, controls, and results for each task are reported in
\Cref{app:causal}.

\section{Why Code Forms Diverge}
\label{sec_form_divergence}

The previous section established Brewing as a recurring interval between
Availability and Readiness and showed that the states on either side have
different functional consequences. We now use this coordinate to examine why
semantically equivalent code forms can produce different answers. We first
locate the divergence between matched Loop and Unrolled programs. We then
identify the late computation associated with this divergence in the anchor
model and examine whether the same coordinates distinguish other code
primitives.

\subsection{Divergence within Brewing}
\label{subsec_brewing_divergence}

We compare Loop and Unrolled programs that execute the same updates from the
same initial values and return the same answer. The difference in FPCL is small,
indicating that the two forms reach Availability at comparable points. Their
trajectories separate after Availability has been reached. Across the matched
comparison, Loop samples more often proceed from FPCL to FJC, while Unrolled
samples more often fail to complete Brewing within the available model depth.
The paired diagnostic therefore places the divergence in whether an available
answer reaches Readiness. Analyses across controlled program
conditions support the same localization and appear in \Cref{app:loop-comparison}.

Having located the divergence within Brewing, we next examine the late
computation that produces this separation in the anchor model.

\subsection{Late MLP Consolidation}
\label{subsec_late_mlp}

To identify this computation, we focus on the dual-variable condition, where
each update changes two coupled state variables. This condition provides 107
matched Loop and Unrolled pairs for component analysis. We conduct the analysis
on Qwen2.5-Coder-7B, the anchor model used throughout the matched comparison.

The two forms remain closely aligned through layer 21. From layer 22 onward,
the Loop representation increasingly favors the correct answer relative to the
matched Unrolled representation. The Loop-minus-Unrolled correct-answer logit
difference reaches 2.76 at layer 26. We decompose each residual update over layers 22 through 27 into its
attention and MLP contributions. Across this segment, MLP updates contribute
1.34 to the Loop advantage, while attention updates offset the advantage by
0.97. At layer 26, the MLP contribution is 1.35, while attention offsets it by
only 0.05.

To test whether the late MLP write drives the difference, we transplant the
Loop twin's MLP outputs at layers 25 and 26 into the matched Unrolled run.
At the logit level, the intervention closes 82.6\% of the correct-answer logit
gap across 69 pairs satisfying the gap criterion. At the output level, it
corrects 13 of 70 initially incorrect Unrolled outputs (18.6\%). Transferring attention at layer 26 closes 10.9\% of the
gap and changes no outputs. Transferring an MLP output from a random donor
closes 29.5\% and also changes no outputs. The matched transfer therefore
identifies the late MLP write carried by the corresponding Loop trajectory as
the source of the observed advantage.

Together, the decomposition and transfer results localize the Loop advantage
to stronger late MLP consolidation in this setting. We limit this attribution
to Qwen2.5-Coder-7B and the dual-variable condition. \Cref{app:overprocessed-rewrite}
reports how component attribution for related late-layer effects varies across
model families.

The Loop comparison shows that the paired diagnostic can locate a divergence
at both the stage and component levels. We next ask whether the same stage
coordinates distinguish the bottlenecks produced by other code primitives.

\subsection{Bottlenecks across Code Primitives}
\label{subsec_primitive_bottlenecks}

\begin{figure}[t]
  \centering
  \includegraphics[width=0.85\linewidth]{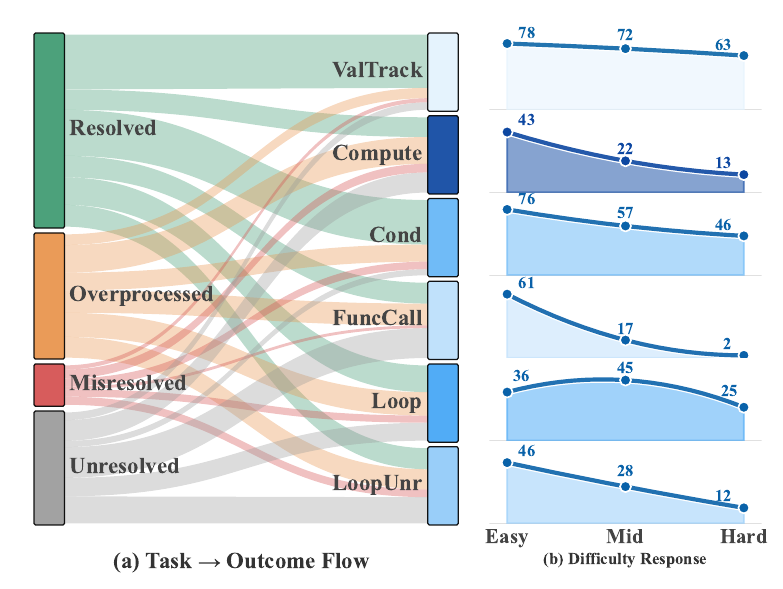}
  \caption{Outcome fingerprints and difficulty responses for
  Qwen2.5-Coder-7B. \textbf{(a)} Sankey diagram linking the six
  task families to the four resolution outcomes: Resolved,
  Overprocessed, Misresolved, and Unresolved. Ribbon width is
  proportional to the sample fraction within each task.
  \textbf{(b)} Resolved rate across three levels of each task's
  primary difficulty dimension: depth for Value Tracking,
  Conditional, and Function Call; arithmetic steps for Computing;
  and iteration count for Loop and Loop-unrolled. Together, the
  panels show that comparable declines in Resolved rate can
  redistribute trajectories across different failure outcomes,
  revealing distinct bottlenecks after Availability in reaching or
  preserving Readiness.}
  \label{fig:fingerprint}
\end{figure}

Computing and Function Call provide a direct contrast because their dominant
failures occur at different stages.

In Computing, increasing the number of arithmetic steps from two to four
reduces Resolved from 42.6\% to 12.6\% and raises Overprocessed from 25.4\% to
47.5\%. FPCL changes little across the same comparison. These trajectories
often make the answer available and reach FJC, yet later processing fails to
preserve the jointly correct state. The dominant bottleneck therefore lies
after Brewing has completed.

Function Call produces a different pattern. Increasing call depth from one to
three reduces Resolved from 61.1\% to 2.5\% and raises Unresolved from 19.1\%
to 57.4\%. FPCL remains comparable, while the share of samples reaching FJC
falls from approximately 79\% to 40\%. The answer continues to become
available, but fewer trajectories complete Brewing and reach Readiness. The
dominant bottleneck therefore lies within Brewing.

These examples show what the paired diagnostic adds to final accuracy. Similar
performance declines can arise because Brewing does not complete or because
later layers alter a state that has already reached Readiness. By locating each
failure relative to Availability and Readiness, the framework makes these
distinct internal bottlenecks directly comparable across code forms and
primitives. Detailed cross-task analyses appear in \Cref{app:bottleneck}.

\section{Conclusion}

Final accuracy collapses distinct internal trajectories into the same output label.
Our study shows that a correct answer can differ in whether it is already
represented, whether it has become usable by the model's own computation, and
whether that state is preserved until the final prediction. Separating these stages
with paired diagnostics lets us locate where reasoning trajectories diverge, not
only observe where they fail.

Using this view, the framework traces a behavioral difference to a specific point
in internal computation. For matched Loop and Unrolled programs, the divergence lies
in the transition from Availability to Readiness. On the anchor model, this difference further localizes to late MLP consolidation in the dual-variable condition. The framework thus connects a difference in outputs to a specific internal computation that final accuracy alone
would leave hidden.

Beyond this case, the same diagnostics distinguish different failure patterns across
reasoning primitives. Some trajectories fail because the transition toward Readiness
does not complete. Others fail once later processing overwrites a state that had
already reached it. These distinctions suggest that improving reasoning systems
requires understanding when additional computation helps and when later processing
alters a state that had already reached Readiness. The framework therefore provides
a basis for analyzing and designing adaptive computation strategies that account for
internal reasoning trajectories.
\section*{Limitations}

Our experiments focus on short, deterministic Python programs with answers restricted
to single-digit tokens. This controlled setting enables consistent layer-wise
comparison across tasks and models, but it does not capture the autoregressive
dependencies of multi-token generation or the compositional complexity of longer
programs. The CRUXEval-O evaluation provides complementary evidence that model-level
capability trends extend to a real execution benchmark, while the full diagnostic
analysis remains limited to the controlled setting. Extending the framework to
richer outputs, longer programs, and realistic software engineering tasks remains an
important direction.

A second limitation concerns the interpretation of the proposed diagnostics.
Availability and Readiness are operational definitions based on linear probing and
Context-Stripped Decoding. Like other interpretability methods, they measure internal
computation through specific readout conditions and interventions, and therefore do
not capture every possible form in which information may be represented or used.
The framework should be viewed as a set of diagnostic coordinates for locating
internal transitions rather than a complete description of reasoning processes.
Future work can examine alternative representations, broader reasoning domains, and
additional forms of internal computation.

\bibliography{custom}

\appendix
\ifarxiv
  \newpage
  \appendixcompact
  \begin{center}
    {\Large\bfseries\color{cuebluedark} Appendix}
  \end{center}
  \vspace{1em}
\else
  \twocolumn[%
    \begin{center}
      \Large\bfseries Appendix
    \end{center}
    \vspace{1em}%
  ]
\fi
\section{Experimental Setup and CUE-Bench Construction}
\label{app:setup}
\label{app_experimental_setup}

The paired diagnostic requires layer-wise comparisons in which the answer space, prompt format, and code execution target remain fixed. We therefore use CUE-Bench as a controlled setting for tracing Availability, Readiness, and Brewing. The benchmark contains short deterministic Python programs with execution-verified answers and systematic variation over code structure and difficulty. This design supports matched hidden-state analysis and intervention while keeping the scope limited to controlled code reasoning. Long-form code generation, repository-level interaction, and software engineering workflows remain outside the present diagnostic setting.

\subsection{Model Configuration and Inference Setup}
\label{app:models}
\label{app:model-config}
\label{app_model_configuration}

We evaluate 16 decoder-only Transformer models spanning the Qwen2.5-Coder, Qwen2.5, Qwen3, DeepSeek-Coder, CodeLlama, and Llama-2 families. The evaluated scales range from 0.5B to 14B parameters. Qwen2.5-Coder-7B serves as the anchor model for the matched Loop comparison and all functional and component analyses. Cross-task outcome analyses also use this model unless another scope is stated explicitly. Cross-model analyses use the complete model set. \Cref{app:model-inventory} reports the model identifiers, parameter scales, and evaluation roles.

All models are evaluated with their original pretrained weights. Inference uses \texttt{bf16} precision without additional fine-tuning, quantization, or instruction tuning. Final predictions are obtained through greedy next-token decoding from the shared prompt format described below. Each model is evaluated on 4,050 instances from each task family, giving 24,300 evaluation instances across the six families.

Probe fitting uses a separately generated split, while all landmarks, outcome labels, interventions, and reported task statistics are computed on held-out evaluation instances. All hidden states from one program remain in the same split. This instance-level separation prevents a probe from observing another layer of an evaluation program during training. \Cref{app:probe} gives the hidden-state extraction and classifier details.

\subsection{CUE-Bench Construction}
\label{app:bench}
\label{app:cue-bench}
\label{app_cue_construction}

CUE-Bench isolates code computations that can be varied without changing the output format or the diagnostic target. Each instance is a short deterministic Python program whose answer can be checked by execution. The six task families cover value propagation, arithmetic computation, branch selection, computation across function boundaries, and repeated state updates. \Cref{tab_cue_task_families} summarizes the computation isolated by each family.

\begin{table}[H]
\centering
\small
\resizebox{\columnwidth}{!}{%
\begin{tabular}{p{0.16\textwidth} p{0.37\textwidth} p{0.39\textwidth}}
\toprule
Task family & Controlled computation & Program forms \\
\midrule
Value Tracking & Propagation of an input value without changing it & Function chains, containers, and method chains with optional distractors \\
Computing & Multi-step arithmetic that produces a single final value & Function arithmetic, chained calls, and accumulator programs \\
Conditional & Selection of the executed branch for a fixed input & \texttt{if}/\texttt{elif} chains, guard clauses, and sequential \texttt{if} statements \\
Function Call & Value propagation combined with computation across nested function boundaries & Arithmetic functions, container relay, and conditional returns \\
Loop & Repeated updates to one or two state variables & Simple accumulation, filtered counting, and dual-variable updates expressed with a \texttt{for} loop \\
Loop-unrolled & The same repeated updates expressed as explicit sequential statements & Matched unrolled forms of the Loop templates \\
\bottomrule
\end{tabular}%
}
\caption{The six CUE-Bench task families and the computation isolated by each family.}
\label{tab_cue_task_families}
\end{table}

Each task family contains three dimensions with three levels each, yielding 27 configurations. The generator targets 150 instances for every configuration, which gives 4,050 instances per family. For each candidate, the generator samples program parameters and identifiers, executes the program in a restricted Python environment, and retains the instance only when execution succeeds and the answer lies in the target digit set. The saved record contains the program, the result variable, the verified answer, and the three configuration values used for later difficulty analysis.

All task families use the same completion format.

\begin{quote}
\small\ttfamily
\{code\}\\
\# The value of \{variable\} is "
\end{quote}

The prompt ends immediately before the answer token. The hidden state at this final prompt position can attend to the complete program and provides a consistent extraction point across task families and model depths.

\subsection{Representative Code Examples}
\label{app:code-examples}

To illustrate the controlled computations evaluated in CUE-Bench, we provide
representative examples from three task families. These examples are selected
to highlight the structural differences analyzed in the main text. The
generated benchmark contains additional variations over difficulty dimensions,
identifier choices, and irrelevant context, while preserving the same answer
interface.

\subsubsection{Loop and Loop-unrolled}

Loop and Loop-unrolled form a matched comparison in which the executed
computation is identical while the code structure differs. The following pair
uses the same dual-variable update with four iterations and the same verified
answer.

\paragraph{Loop.}

{\small
\begin{verbatim}
def step(n, start_a, start_b):
    a = start_a
    b = start_b
    for i in range(n):
        a, b = b, a + 1
    return a

result = step(4, 3, 4)
# The value of result is "
\end{verbatim}
}

The model must track the repeated update sequence:
$(3,4) \rightarrow (4,4) \rightarrow (4,5)
\rightarrow (5,5) \rightarrow (5,6)$.
The verified answer is 5.

\paragraph{Loop-unrolled.}

{\small
\begin{verbatim}
def step():
    a = 3
    b = 4
    a, b = b, a + 1
    a, b = b, a + 1
    a, b = b, a + 1
    a, b = b, a + 1
    return a

result = step()
# The value of result is "
\end{verbatim}
}

The unrolled version preserves the same state transitions without an explicit
loop structure. The answer remains 5, allowing the comparison to isolate the
effect of code form.

\subsubsection{Function Call}

Function Call evaluates value propagation when computation is performed across
nested function boundaries. The following example uses three nested functions
with conditional computation.

{\small
\begin{verbatim}
val = 2

def route(x):
    if x > 3:
        return x + 2
    return x - 1

def dispatch(y):
    tmp = route(y)
    return tmp + 1

def process(z):
    mid = dispatch(z)
    return mid + 1

result = process(val)
# The value of result is "
\end{verbatim}
}

The model must resolve each function boundary and propagate the intermediate
values through the complete call chain. The verified answer is 3.

\subsubsection{Conditional}

Conditional isolates control-flow reasoning by requiring the model to determine
which execution branch is selected.

{\small
\begin{verbatim}
def classify(x):
    if x >= 42:
        return 7
    else:
        return 3

result = classify(55)
# The value of result is "
\end{verbatim}
}

Because the input satisfies the branch condition, the executed path returns 7.
More complex Conditional configurations vary branch structure, nesting depth,
and predicate form while keeping the output interface unchanged.

\subsection{Structural Controls and Difficulty Dimensions}
\label{app:structural-controls}
\label{app:task-spectrum}
\label{app_structural_controls}

The benchmark includes controlled comparisons that separate code form from the computation being executed. Value Tracking and Function Call use related propagation structures, while Function Call applies computation at each function boundary. This comparison tests how added computation across boundaries changes the trajectory relative to direct value propagation. Loop and Loop-unrolled provide the stricter matched comparison. Their initial values, update operations, number of repetitions, and verified answer are held fixed, while the repeated computation is written with loop syntax or as sequential statements. \Cref{sec_form_divergence} and \Cref{app:loop-mechanism} use this comparison to locate where the two trajectories separate.

\Cref{tab_cue_difficulty_dimensions} lists the dimensions used to vary each task family. The dimensions change program structure, computation length, or irrelevant context while retaining the same answer interface. Difficulty analyses vary one dimension and marginalize over the remaining two. This design shows whether a lower Resolved rate is accompanied by incomplete transition toward Readiness, preservation failure after Readiness, or another outcome shift.

\begin{table*}[tp]
\centering
\scriptsize
\resizebox{\textwidth}{!}{%
\begin{tabular}{llll}
\toprule
Task family & First dimension & Second dimension & Third dimension \\
\midrule
Value Tracking & mechanism \{\texttt{function\_chain}, \texttt{container}, \texttt{method\_chain}\} & depth \{1, 2, 3\} & distractors \{0, 1, 2\} \\
Computing & structure \{\texttt{func\_arithmetic}, \texttt{chained\_calls}, \texttt{accumulator}\} & steps \{2, 3, 4\} & operators \{\texttt{add}, \texttt{add\_sub}, \texttt{add\_mul}\} \\
Conditional & branch form \{\texttt{elif\_chain}, \texttt{guard\_clause}, \texttt{sequential\_if}\} & depth \{1, 2, 3\} & condition form \{\texttt{numeric}, \texttt{membership}, \texttt{boolean\_flag}\} \\
Function Call & mechanism \{\texttt{arithmetic}, \texttt{container\_relay}, \texttt{conditional\_return}\} & depth \{1, 2, 3\} & distractors \{0, 1, 2\} \\
Loop & body form \{\texttt{simple\_acc}, \texttt{filter\_count}, \texttt{dual\_var}\} & iterations \{2, 3, 4\} & initial offset \{\texttt{zero}, \texttt{low}, \texttt{high}\} \\
Loop-unrolled & body form \{\texttt{simple\_acc}, \texttt{filter\_count}, \texttt{dual\_var}\} & iterations \{2, 3, 4\} & initial offset \{\texttt{zero}, \texttt{low}, \texttt{high}\} \\
\bottomrule
\end{tabular}%
}
\caption{Difficulty dimensions for the six task families. Every combination of the three dimensions defines one configuration.}
\label{tab_cue_difficulty_dimensions}
\end{table*}

Identifiers are randomized from shared pools of function, variable, parameter, class, and method names. A generated program does not reuse an identifier within the same snippet. Distractor parameters and fields are sampled from the same pools and do not affect the executed answer. These controls reduce reliance on repeated lexical cues while preserving the code structure required by each configuration.

\subsection{Answer Space Validation}
\label{app:answer-validation}
\label{app:prompt-format}
\label{app:answer-space}
\label{app:difficulty}
\label{app_answer_space}

The digit answer space is a measurement control for the paired diagnostic. Every ground-truth answer belongs to $\mathcal{D}=\{0,\ldots,9\}$ and is represented by one digit token in the evaluated models. This common target removes variation in answer length and token boundaries from layer-wise comparison. Probing extends the ten answer classes with the near-OOD residual class described in \Cref{app:probe}. The resulting claims concern answer formation under this controlled single-token setting. Multi-token generation remains outside the full diagnostic analysis.

\begin{figure}[H]
\centering
\includegraphics[width=\columnwidth]{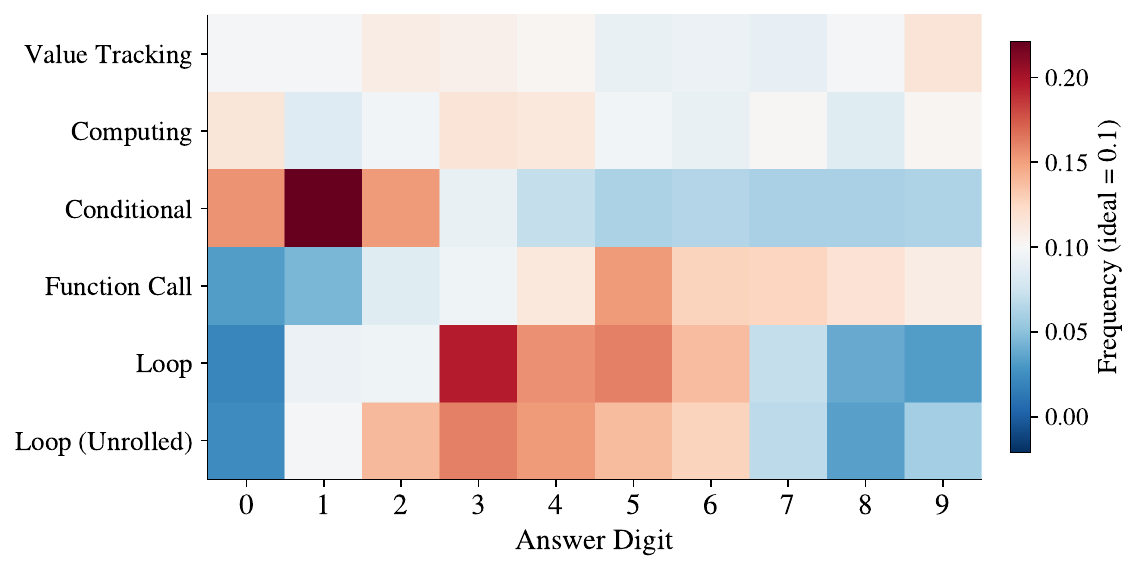}
\caption{Answer frequencies over digits 0--9 for the six CUE-Bench task families. All diagnostics use per-instance ground truth, and \Cref{app:csd} removes fixed target-prompt preferences through CSD baseline subtraction.}
\label{fig_answer_distribution}
\label{fig:answer-dist}
\end{figure}

Generation uses rejection sampling to retain only programs whose executed result lies in $\mathcal{D}$. Operands and initial values are constrained when necessary to keep the result within this range, and every accepted program is executed again before it is saved. The generators do not force a uniform answer distribution. \Cref{fig_answer_distribution} reports the observed frequency of each digit for every task family.

The observed skew is therefore treated as part of each task distribution rather than as evidence for a uniform benchmark. Probes are fitted and evaluated within each task family, outcome labels use the exact executed answer for every instance, and CSD subtracts the target-prompt baseline before comparing digit scores. These controls prevent a fixed digit preference from being counted as Availability or Readiness.

We also use a limited two-digit pilot as a scope check. The pilot contains 200 samples drawn across the six task families and uses Qwen2.5-Coder-7B. FPCL precedes FJC for the tens digit. Brewing for the ones digit remains visible when the tens digit is correct, while an incorrect tens digit perturbs the later trajectory and replacing it with the correct prefix improves ones-digit CSD. This result indicates that adding a second answer position does not remove the separation between Availability and Readiness in the tested pilot. Its limited size and dependence on the generated prefix do not support extending the main quantitative claims beyond the single-token setting.

\section{Cross-Model Scaling and External Validation}
\label{app:crossmodel}
\label{app:scaling}

The previous sections analyze internal trajectories primarily through the anchor model. We next examine whether the observed separation between Availability and Readiness extends across model scales, training settings, and architectures. These analyses are intended to characterize the consistency of the paired diagnostic framework across models rather than to establish a universal scaling rule.

\subsection{Coder vs.\ Base Comparison}
\label{app:coder-base}

Code-specialized models provide a controlled comparison for examining how training data affects resolution outcomes. We compare matched Qwen2.5-Coder and Qwen2.5 Base models at the same parameter scales while keeping the evaluation tasks and diagnostic procedure unchanged.

\begin{figure}[H]
\centering
\includegraphics[width=\columnwidth]{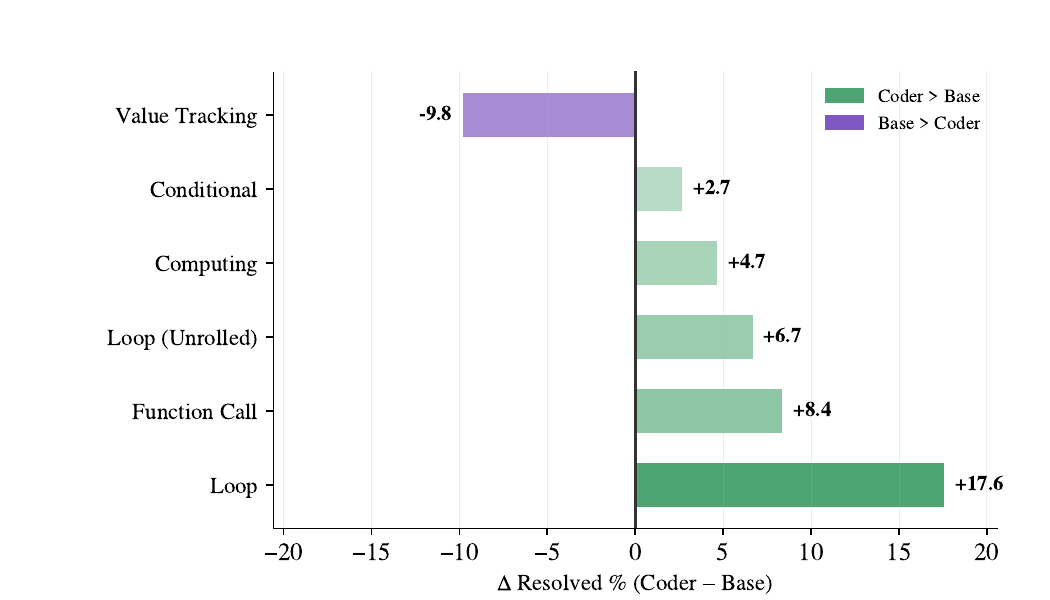}
\caption{Outcome differences between Qwen2.5-Coder and matched Qwen2.5 Base models. Code-specialized training shifts trajectories toward Resolved outcomes while retaining Overprocessed and Unresolved trajectories.}
\label{fig:coder-base}
\end{figure}

\Cref{fig:coder-base} shows the difference in Resolved rate between matched Coder and Base models across task families. Code-specialized training generally increases the fraction of trajectories that reach successful resolution, but the change is task dependent. Loop and Function Call show the largest improvements, while Value Tracking changes less and Conditional shows a smaller shift.

The comparison also shows that improved final performance does not remove the internal separation measured by the paired diagnostics. Code-specialized models achieve more Resolved trajectories, but Overprocessed and Unresolved trajectories remain present. The result suggests that training changes the frequency with which trajectories resolve while preserving the distinction between different failure outcomes.

\subsection{Scaling Trends}
\label{app:scaling-trends}

We next analyze how resolution outcomes change with model scale. We evaluate the Qwen2.5-Coder series from 0.5B to 14B parameters and examine outcome distributions across the six task families.

\begin{figure}[H]
\centering
\includegraphics[width=\columnwidth]{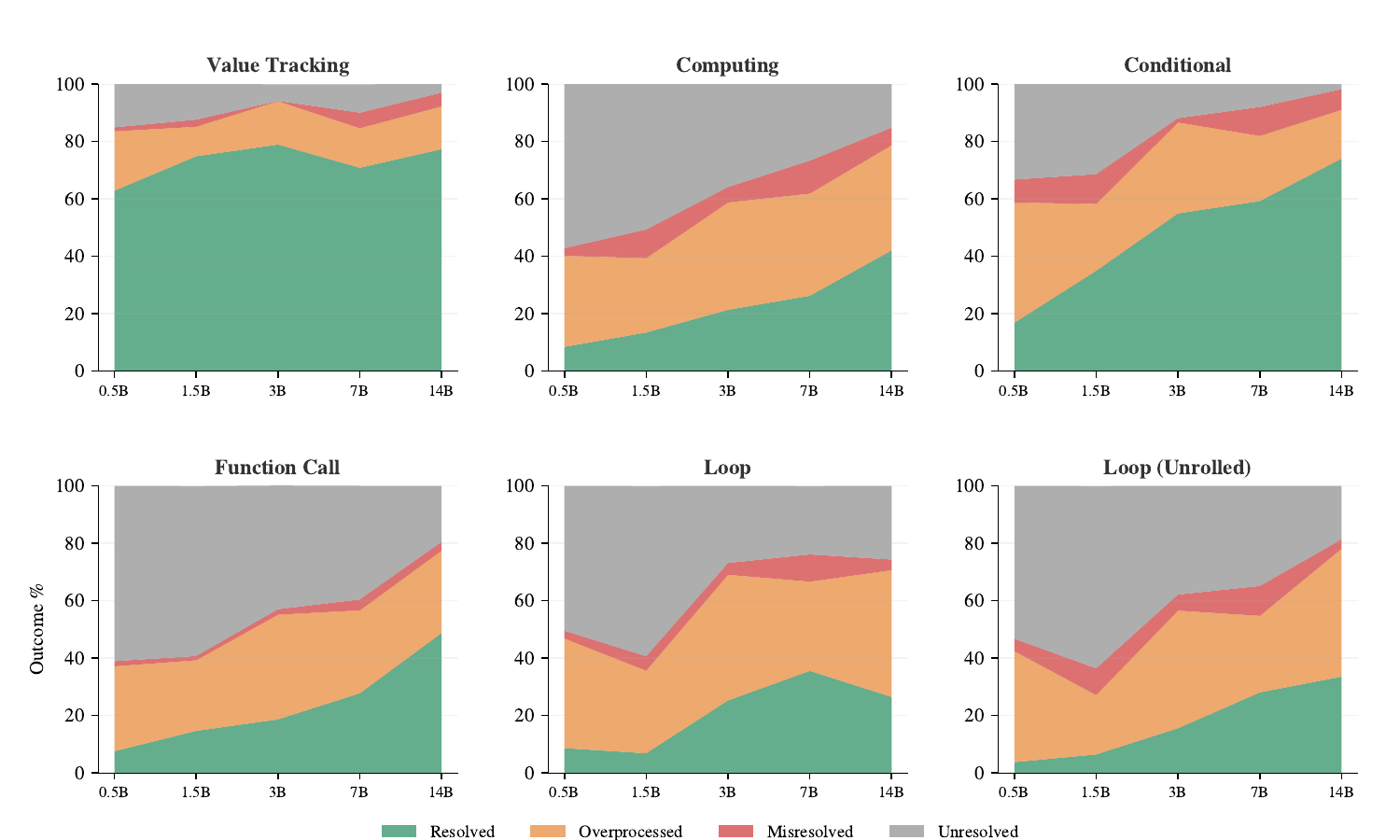}
\caption{Resolution outcome proportions across Qwen2.5-Coder model scales.}
\label{fig:outcome-scaling}
\end{figure}

\Cref{fig:outcome-scaling} shows that increasing model scale primarily changes the relative frequency of outcomes. Larger models generally produce more Resolved trajectories and fewer Unresolved trajectories across task families. However, Overprocessed trajectories remain substantial even at larger scales, indicating that higher capability does not simply remove all failure patterns identified by the framework.

We further examine the normalized locations of FPCL and FJC across scales. Larger models tend to reach FPCL at relatively earlier normalized depths, indicating that answer information becomes recoverable earlier in the network. The transition toward Readiness remains separated from this earlier onset.

Across the evaluated Qwen2.5-Coder models, normalized Brewing duration remains within the range reported in the main text. This trend is descriptive and does not imply a fixed scaling relationship. It indicates that the paired diagnostic separation is not specific to a single model size.

\subsection{Cross-Architecture Robustness}
\label{app:cross-arch}

We extend the analysis beyond the Qwen2.5 family by evaluating additional decoder-only Transformer models from different model families.

\begin{figure}[H]
\centering
\includegraphics[width=0.85\columnwidth]{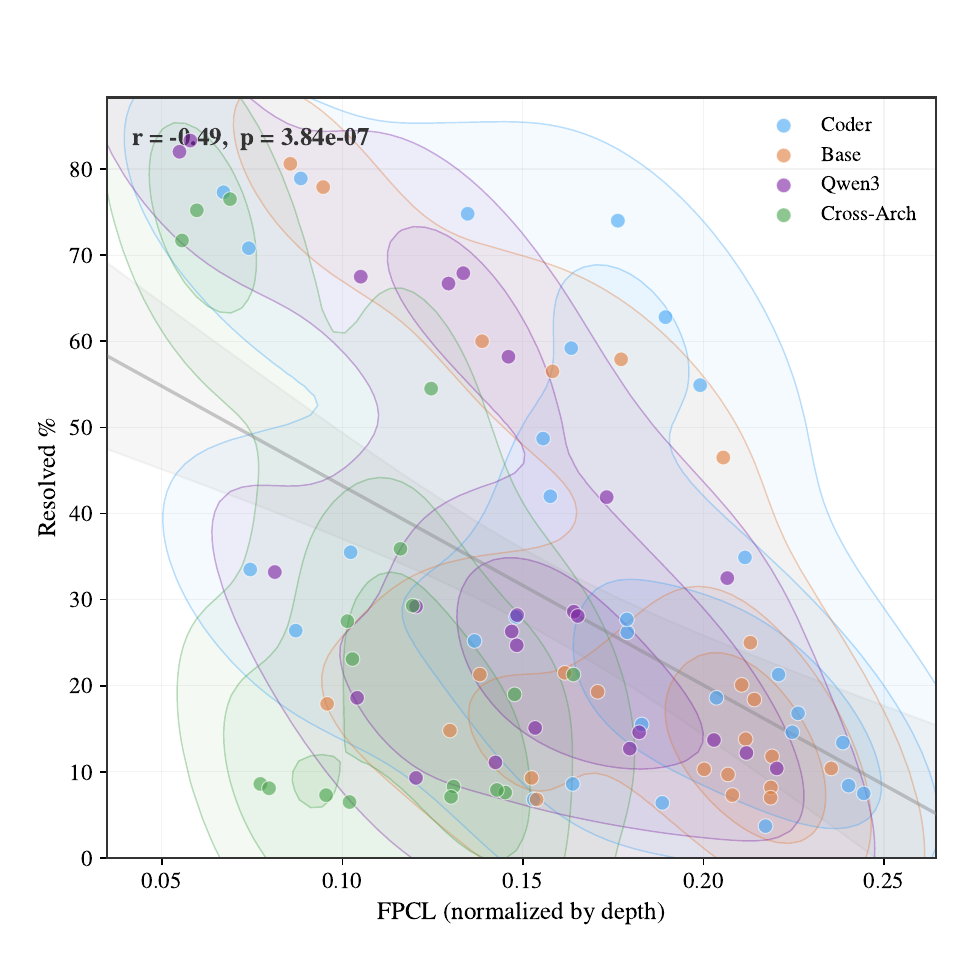}
\caption{Normalized FPCL locations across model scales and architectures.}
\label{fig:fpcl-cloud}
\end{figure}

The evaluated models include Qwen, Llama, DeepSeek-Coder, and CodeLlama variants. \Cref{fig:fpcl-cloud} summarizes the normalized FPCL distribution across scaling and architecture comparisons.

\begin{figure}[H]
\centering
\includegraphics[width=0.75\columnwidth]{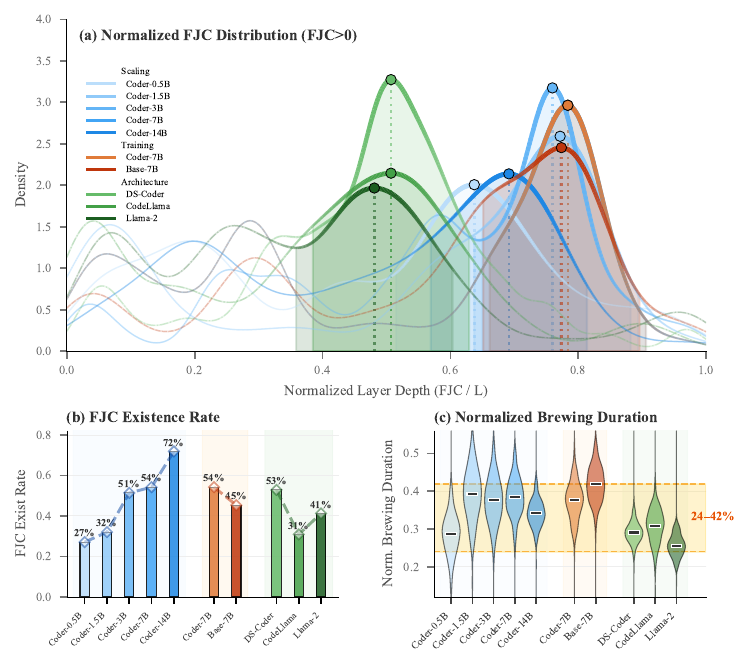}
\caption{Normalized Brewing duration across models and task families. The separation between Availability and Readiness remains observable across the evaluated models.}
\label{fig:brewing-stability}
\end{figure}

Across architectures, the same qualitative pattern appears. FPCL and FJC remain separated, and normalized Brewing duration occupies a similar range across the evaluated models (\Cref{fig:brewing-stability}). The exact outcome proportions vary between model families, reflecting differences in model capability and training.

The cross-architecture comparison also shows that earlier Availability does not directly determine final resolution success. Some models reach FPCL relatively early while still exhibiting substantial unresolved trajectories. Therefore, the paired diagnostics separate early recoverability from successful completion of the remaining computation.

Component-level analyses should be interpreted separately from these trajectory-level comparisons. The component-level interventions reported in \Cref{app:causal} show that the component associated with late-layer changes can vary across model families. The cross-model result supports the presence of similar trajectory stages, while not assigning the same internal component to every model.

\subsection{CRUXEval-O}
\label{app:cruxeval}

CUE-Bench provides controlled execution primitives that allow layer-wise
trajectory analysis. To examine whether the resulting model-level capability
ordering extends beyond this setting, we additionally evaluate models on
CRUXEval-O \citep{gu2024cruxeval}, an execution-based output-prediction
benchmark whose answers are multi-token literals scored by executable
assertions of the form \texttt{assert f(input)==pred}. We use the official
direct few-shot prompt with greedy decoding and
report execution-based pass@1 on a fixed 200-example subset (seed~42) of the
800-example test set. CRUXEval-O is used only for external capability
comparison and does not enter the layer-wise diagnostic pipeline; its output
format is therefore not assigned FPCL, FJC, or a resolution outcome.

\begin{table}[H]
\centering
\footnotesize
\setlength{\tabcolsep}{4pt}
\renewcommand{\arraystretch}{0.88}
\begin{tabular}{@{}lr@{}}
\toprule
Model & Accuracy \\
\midrule
\multicolumn{2}{@{}l}{\emph{Qwen2.5-Coder}} \\
Qwen2.5-Coder-0.5B & 1.0\% \\
Qwen2.5-Coder-1.5B & 40.0\% \\
Qwen2.5-Coder-3B & 39.0\% \\
Qwen2.5-Coder-7B (anchor) & 50.5\% \\
Qwen2.5-Coder-14B & 54.0\% \\
\midrule
\multicolumn{2}{@{}l}{\emph{Qwen2.5 Base}} \\
Qwen2.5-0.5B & 7.0\% \\
Qwen2.5-1.5B & 33.5\% \\
Qwen2.5-3B & 35.0\% \\
Qwen2.5-7B & 49.0\% \\
Qwen2.5-14B & 51.0\% \\
\midrule
\multicolumn{2}{@{}l}{\emph{Qwen3 Base}} \\
Qwen3-0.6B & 27.5\% \\
Qwen3-1.7B & 35.5\% \\
Qwen3-4B & 45.5\% \\
Qwen3-8B & 51.0\% \\
Qwen3-14B & 55.0\% \\
\midrule
\multicolumn{2}{@{}l}{\emph{Cross-architecture (approximately 7B)}} \\
DeepSeek-Coder-6.7B & 45.5\% \\
CodeLlama-7B & 36.5\% \\
Llama-2-7B & 17.5\% \\
\midrule
\multicolumn{2}{@{}l}{\emph{Instruction-tuned}} \\
Qwen2.5-Coder-1.5B-Instruct & 34.0\% \\
Qwen2.5-Coder-0.5B-Instruct & 27.0\% \\
\bottomrule
\end{tabular}
\caption{CRUXEval-O accuracy using execution-based pass@1, greedy decoding,
and the official direct few-shot prompt on a fixed 200-example subset
(seed~42). The table contains the 16 main-study models together with two 14B
base-model extensions and two instruction-tuned variants.}
\label{tab:cruxeval-o}
\end{table}

\Cref{tab:cruxeval-o} shows a broad improvement with scale, although the trend
is not strictly monotonic: Qwen2.5-Coder-1.5B and 3B obtain 40.0\% and 39.0\%,
respectively. At every matched scale of at least 1.5B, the code-specialized
Qwen2.5-Coder model matches or exceeds its Base counterpart. Among the
approximately 7B cross-architecture models, DeepSeek-Coder reaches 45.5\%,
followed by CodeLlama at 36.5\% and Llama~2 at 17.5\%. These comparisons recover
the broad capability ordering observed in CUE-Bench
(\Cref{app:coder-base,app:scaling-trends}) without assuming a fixed scaling law.

The 0.5B pair reveals a separate output-format effect. On the 200-example
subset, Qwen2.5-Coder-0.5B obtains 1.0\%, compared with 7.0\% for the matched
Base model. On the full 800-example set, the corresponding accuracies are
0.5\% and 6.75\%. Inspection of the generated answers shows that the Coder
model copies the prompt's \texttt{??} placeholder on 98.8\% of examples,
compared with 58.1\% for the Base model. Conditional on producing an answer
literal, the reported accuracies are 40.0\% and 16.1\%, respectively, although
this conditional comparison is based on the small subset of outputs that
contain a literal. Instruction tuning raises the Coder-0.5B score to 27.0\% on
the 200-example subset. This pattern is consistent with output formatting
contributing substantially to the raw accuracy gap, but it is not itself a
measurement of Availability or Readiness.

The external evaluation therefore provides complementary evidence that the
model differences observed in CUE-Bench correspond to broader code-execution
capability trends. The detailed trajectory analysis remains restricted to the
controlled setting where hidden states and interventions can be measured.

\subsection{Complete Model Inventory}
\label{app:model-inventory}

\Cref{tab:model-inventory} lists the complete model inventory used in the experiments. We report model family, parameter scale, and evaluation role. The evaluated models are based on publicly released model families and their corresponding technical reports, including Qwen2.5-Coder \citep{hui2024qwen2}, Qwen2.5 \citep{yang2024qwen25}, Qwen3 \citep{yang2025qwen3}, DeepSeek-Coder \citep{guo2024deepseek}, CodeLlama \citep{roziere2023code}, and Llama~2 \citep{touvron2023llama}.

\begin{table}[H]
\centering
\small
\begin{tabular}{llll}
\toprule
Model & Family & Params & Role \\
\midrule
Qwen2.5-Coder-0.5B & Qwen & 0.5B & Scaling \\
Qwen2.5-Coder-1.5B & Qwen & 1.5B & Scaling \\
Qwen2.5-Coder-3B & Qwen & 3B & Scaling \\
Qwen2.5-Coder-7B & Qwen & 7B & Anchor \\
Qwen2.5-Coder-14B & Qwen & 14B & Scaling \\
Qwen2.5 Base & Qwen & -- & Training \\
Qwen3 & Qwen & -- & Architecture \\
DeepSeek-Coder & DeepSeek & -- & Architecture \\
CodeLlama & Llama & -- & Architecture \\
Llama-2 & Llama & -- & Architecture \\
\bottomrule
\end{tabular}
\caption{Complete model inventory and evaluation roles.}
\label{tab:model-inventory}
\end{table}

\section{Diagnostic Framework Details and Reliability Analysis}
\label{app:diagnostic}
\label{app:diagnostic_details}

The paired diagnostic framework relies on two different readout conditions applied to the same hidden state. Probing measures whether the correct answer is linearly recoverable, while Context-Stripped Decoding measures whether the model's remaining computation can decode the transferred state after the original code context has been removed. This section gives the complete implementation of both diagnostics and evaluates whether the resulting FPCL and FJC landmarks are stable enough to support the trajectory analyses in the main text.

\subsection{Linear Probing Implementation}
\label{app:probe}
\label{app:probing}
\label{app:probing_training_details}

For each program, the source prompt is written as $S=[C;Q]$, where $C$ contains the generated Python program and $Q$ is the shared question suffix. For a model with $L$ transformer layers, we extract the final-position hidden state
\(
h^{\ell}\in\mathbb{R}^{d}
\)
after every layer
\(
\ell\in\{0,\ldots,L-1\}.
\)
The final position can attend to the complete source prompt and therefore provides the same extraction location across task families and model depths.

For each model, task family, and layer, we fit an independent multinomial logistic classifier

\begin{equation}
    \Phi^{\ell}_{P}(h^{\ell})
    =
    \operatorname{softmax}
    \left(
        W_{\ell}h^{\ell}+b_{\ell}
    \right),
    \label{eq:appendix_probe}
\end{equation}

where
\(
W_{\ell}\in\mathbb{R}^{11\times d}
\)
and
\(
b_{\ell}\in\mathbb{R}^{11}.
\)
A layer is Probing correct when the classifier assigns its highest probability to the executed answer $t^{*}$.

The classifier operates over
\(
\mathcal{T}=\mathcal{D}\sqcup\{\bar d\},
\)
where
\(
\mathcal{D}=\{0,\ldots,9\}
\)
contains the ten answer digits and $\bar d$ is the residual class. To construct training examples for $\bar d$, we use the same task generators while relaxing the single-digit answer constraint. The resulting programs retain the structure of the corresponding task family, while their executed answers contain multiple digits and therefore fall outside $\mathcal{D}$. The number of residual examples is balanced against the average number of training examples for each digit class.

The residual class gives the classifier an explicit alternative to selecting one of the ten candidate answers. When $\bar d$ receives the highest score, the learned linear readout does not assign the state to a single digit. This reduces early false positives that would otherwise arise because a ten-class classifier must select some digit even when digit-specific answer information is weak.

Probe-training programs are disjoint from the benchmark instances used for the reported trajectories. The training records are divided at the program level, so every layer from one program remains within the same partition. The additional multi-digit programs used for $\bar d$ are also excluded from the benchmark evaluation counts reported in \Cref{app:setup}.

\begin{table}[tp]
\centering
\small
\begin{tabular}{ll}
\toprule
Setting & Value \\
\midrule
Classifier & Multinomial logistic regression \\
Solver & \texttt{lbfgs} \\
Regularization & L2 with $C=1.0$ \\
Maximum iterations & 1000 \\
Number of classes & 11 \\
Readout position & Final prompt position \\
Training unit & One classifier per layer \\
\bottomrule
\end{tabular}
\caption{Linear Probing configuration. The independent classifier at each layer limits the external readout to a linear transformation, so a correct prediction indicates that the answer is recoverable without additional nonlinear computation by the classifier.}
\label{tab:probe_configuration}
\end{table}

The use of a linear classifier is deliberate. A more expressive classifier could combine hidden features through additional computation and move the apparent onset of Availability earlier for reasons introduced by the classifier itself. The linear readout therefore supplies a conservative and consistent operational landmark across models and tasks.

\subsection{Probe Reliability Analysis}
\label{app:probe-reliability}
\label{app:probe_reliability}

FPCL records the first layer at which Probing predicts the correct answer. This first-attainment definition makes early false positives a direct concern. Low accuracy in the earliest layers is expected when the answer has not yet become linearly recoverable, but unstable classifier fitting could produce the same surface pattern. We therefore examine held-out accuracy across depth, optimization convergence, continuous true-class confidence, and bootstrap uncertainty for the resulting landmarks.

Across the 96 model and task combinations, the mean best-layer accuracy is 80.09\%, with a standard deviation of 13.8\% and a range from 44.3\% to 97.9\%. The variation follows task and model difficulty. Value Tracking reaches a cross-model mean of 94.2\%, while Function Call reaches 63.3\%. On Qwen2.5-Coder-7B, best-layer accuracy ranges from 72.2\% on Function Call to 95.7\% on Value Tracking.

The first 15\% of model depth provides the more relevant check for FPCL. Mean accuracy over this region ranges from 0.20 to 0.72 across the 96 combinations, compared with best-layer values from 0.44 to 0.98. The corresponding true-class confidence curves begin near the chance level of approximately 0.1 and rise through the region containing the mean FPCL. This continuous signal is calculated independently of the binary correctness indicator used to define FPCL. Its increase therefore supports interpreting FPCL as the onset of linearly recoverable answer information rather than an isolated correct argmax.

\begin{figure}[H]
\centering
\includegraphics[width=\columnwidth]{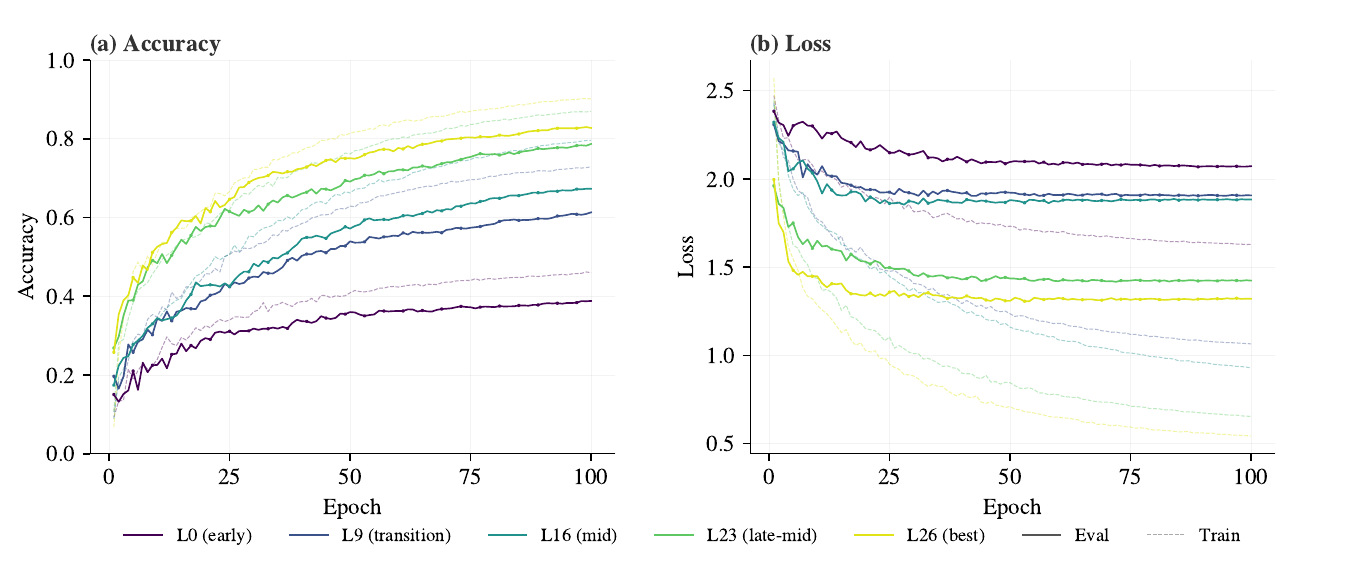}
\caption{Probing optimization curves for representative layers of Qwen2.5-Coder-7B on Computing. Later layers retain higher held-out accuracy after convergence, showing that the layer-wise rise used to locate FPCL follows the hidden states rather than incomplete optimization.}
\label{fig:probe_training_curves}
\label{fig:probe-training}
\end{figure}

We separately expose the optimization process by training equivalent linear classifiers with incremental SGD for 200 epochs on the Computing task of Qwen2.5-Coder-7B. \Cref{fig:probe_training_curves} reports representative layers. Accuracy and loss stabilize without a widening gap between the training and evaluation curves. Deeper layers reach higher held-out accuracy, while the earliest layer remains low after optimization has converged.

\begin{figure}[H]
\centering
\includegraphics[width=\columnwidth]{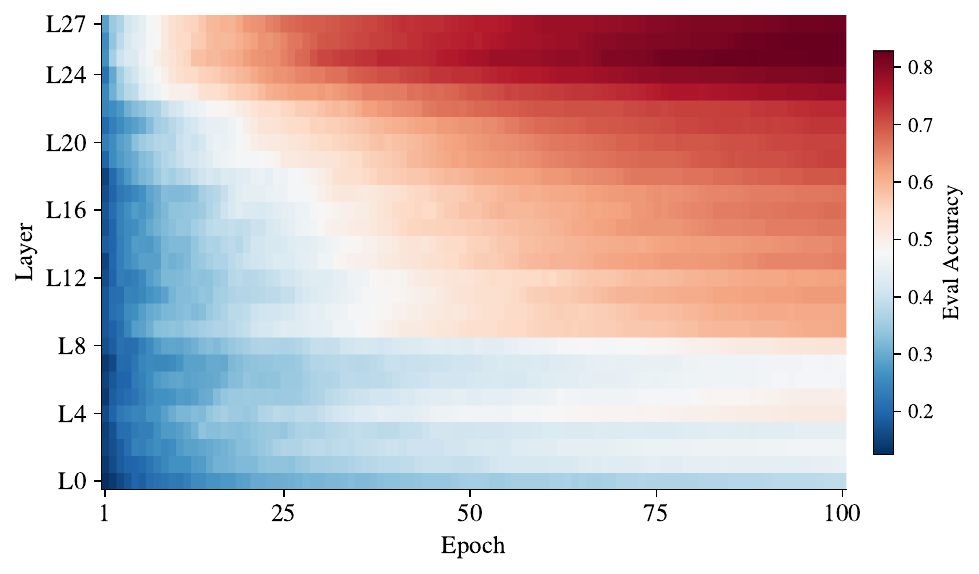}
\caption{Held-out Probing accuracy across all layers and training epochs for Qwen2.5-Coder-7B on Computing. The heatmap verifies that the depth profile used by FPCL remains stable as classifier training proceeds.}
\label{fig:probe_training_heatmap}
\label{fig:diag-heatmap}
\end{figure}

\Cref{fig:probe_training_heatmap} extends the same check to all 28 layers. The deepest layers stabilize after approximately 25 epochs, and shallow layers remain consistently less accurate throughout training. The absence of a late training-driven increase in the early layers reduces the likelihood that FPCL is produced by uneven convergence across depth.

We quantify uncertainty in FPCL, FJC, and Brewing duration through instance-level bootstrap resampling. For every task, we draw 10,000 resamples with replacement and compute 95\% percentile confidence intervals using a fixed random seed. Layer indices are divided by the corresponding model depth before pooling across models. The normalized FPCL and FJC confidence intervals have half-widths between 0.002 and 0.005 at the pooled level.

\Cref{tab:bootstrap_brewing} reports the normalized task-level results. FPCL and FJC are marginal means over trajectories where the corresponding landmark exists. Brewing duration is computed over the paired subset where both landmarks exist, so its mean need not equal the difference between the two marginal means.

\begin{table*}[tp]
\centering
\small
\begin{tabular}{lcccc}
\toprule
Task family &
$\operatorname{FPCL}/L$ &
$\operatorname{FJC}/L$ &
$\Delta_{\mathrm{brew}}/L$ &
95\% CI \\
\midrule
Value Tracking & 0.100 & 0.522 & 0.430 & [0.425, 0.435] \\
Computing & 0.190 & 0.450 & 0.301 & [0.294, 0.308] \\
Conditional & 0.172 & 0.496 & 0.341 & [0.335, 0.346] \\
Function Call & 0.185 & 0.484 & 0.333 & [0.326, 0.340] \\
Loop & 0.118 & 0.420 & 0.329 & [0.322, 0.335] \\
Loop-unrolled & 0.158 & 0.434 & 0.312 & [0.306, 0.319] \\
\bottomrule
\end{tabular}
\caption{Bootstrapped normalized landmarks and Brewing durations pooled over the 16 evaluated models. Narrow intervals show that the FPCL--FJC separation is not dominated by sampling variation.}
\label{tab:bootstrap_brewing}
\end{table*}

These checks do not require every trajectory-level FPCL to be error free. They establish that the depth profile is stable after classifier convergence, that true-class confidence rises around the identified onset, and that the aggregate landmarks used by the paper have narrow uncertainty intervals. The subsequent outcome robustness and intervention analyses provide separate checks on conclusions that depend on these landmarks.

\subsection{Context-Stripped Decoding Implementation}
\label{app:csd}
\label{app:context-stripped}
\label{app:subsec:csd_implementation}

CSD evaluates the same final-position state used by Probing under the model's own remaining computation. The source prompt is
\(
S=[C;Q],
\)
while the target prompt is
\(
S_{\mathrm{tgt}}=Q.
\)
The target prompt retains the question suffix and removes the program. A correct answer decoded after transfer must therefore be supported by the transferred state and the remaining model layers under this stripped context.

For each layer $\ell$, we first run the complete source prompt and cache the final-position state
\(
h^{\ell}_{S}.
\)
We then perform a separate target run on $S_{\mathrm{tgt}}$. After target layer $\ell$, we replace its final-position state with the cached source state

\begin{equation}
    \widetilde{h}^{\ell}_{\mathrm{tgt}}
    \leftarrow
    h^{\ell}_{S}.
    \label{eq:csd_state_transfer}
\end{equation}

All other target positions remain unchanged. The model continues from layer $\ell+1$ through layer $L-1$, followed by the final layer normalization and unembedding. Writing $F_j$ for transformer block $j$, $\operatorname{LN}$ for the final normalization, and $W_u$ for the unembedding matrix, the patched logits are

\begin{equation}
    z^{\ell}_{\mathrm{patch}}
    =
    W_u
    \circ
    \operatorname{LN}
    \circ
    F_{L-1}
    \circ
    \cdots
    \circ
    F_{\ell+1}
    \left(
        \widetilde{h}^{\ell}_{\mathrm{tgt}}
        \mid
        S_{\mathrm{tgt}}
    \right).
    \label{eq:csd_patched_logits}
\end{equation}

The notation after the conditioning bar indicates that all remaining attention operations use the target run, which contains only $S_{\mathrm{tgt}}$. At the final transformer layer, the continuation consists of the final normalization and unembedding. Every patched layer is evaluated in an independent target run, so transferred states from different depths do not interact.

The question suffix can create a nonuniform digit preference even when no source state is transferred. We therefore perform one unmodified target run to obtain full-vocabulary baseline logits
\(
z_{\mathrm{base}}.
\)
The corrected logits for layer $\ell$ are

\begin{equation}
    z^{\ell}_{\mathrm{CSD}}
    =
    z^{\ell}_{\mathrm{patch}}
    -
    z_{\mathrm{base}}.
    \label{eq:csd_corrected_logits}
\end{equation}

Subtraction is applied before normalization. The ten digit entries use the corrected logits of the corresponding digit tokens. The residual entry $\bar d$ aggregates the corrected scores assigned to all remaining vocabulary tokens. This produces the same semantic partition used by Probing, with ten candidate digits and one residual alternative.

The CSD distribution is then

\begin{equation}
    \Phi^{\ell}_{C}(h^{\ell})
    =
    \operatorname{softmax}_{\mathcal{T}}
    \left(
        z^{\ell}_{\mathrm{CSD}}
    \right).
    \label{eq:appendix_csd}
\end{equation}

A layer is CSD correct when the highest probability under
\(
\Phi^{\ell}_{C}
\)
is assigned to $t^{*}$. FJC additionally requires Probing correctness at the same layer, which keeps Readiness tied to a state where the answer is also linearly recoverable.

The baseline depends only on the target prompt and model. It is computed once for each program and reused across the $L$ patched runs, giving $L+1$ target forward passes rather than two forward passes per layer.
\begin{figure*}[t]
\centering
\includegraphics[width=\textwidth]{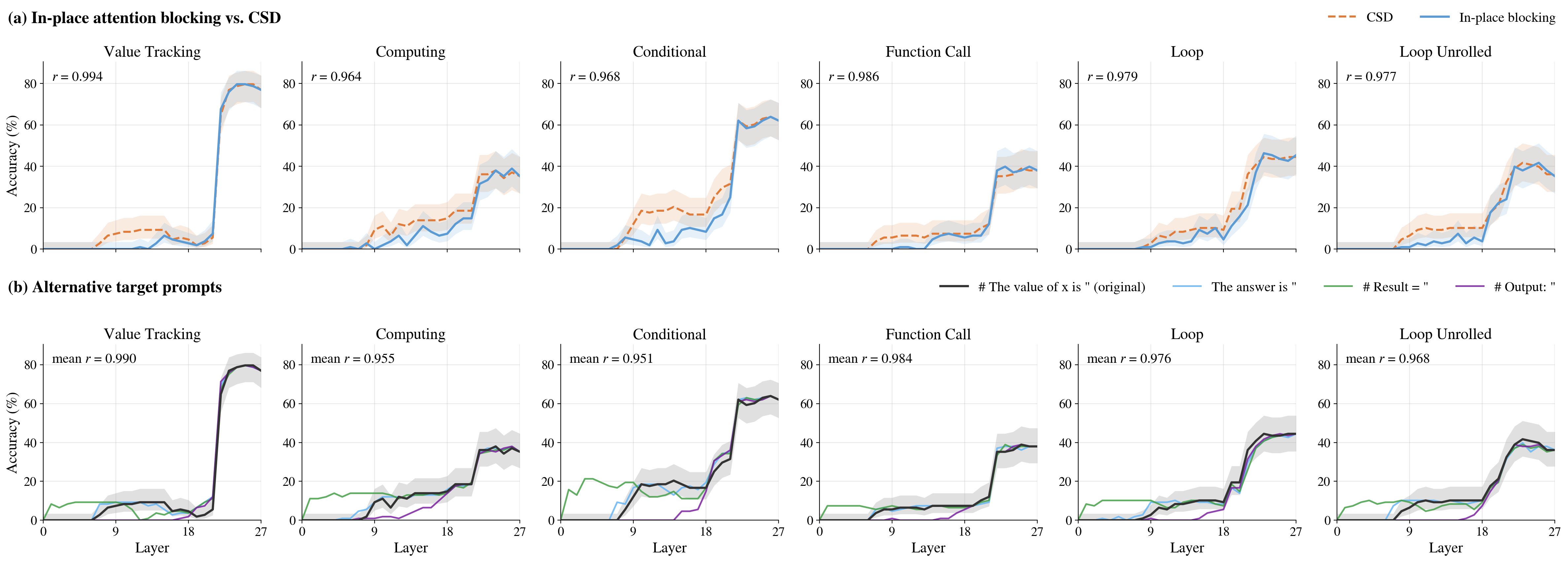}
\caption{CSD validity checks on Qwen2.5-Coder-7B across the six task families. \textbf{(a)} Layer-wise accuracy under CSD and under in-place attention blocking, which preserves the original prompt and blocks attention from question positions to code positions after each cut layer. \textbf{(b)} Layer-wise CSD accuracy under the original target prompt and three alternative target prompts with the same source hidden states. Each panel reports the Pearson correlation with the original CSD curve; in (b), $r$ is averaged over the three alternative prompts.}
\label{fig:csd-validity}
\end{figure*}
We do not apply an additional transformation or positional correction to the transferred source vector. The remaining layers operate with the positions and attention context of the target run. This implementation keeps CSD tied to the model's existing computation while removing access to the original code.

\subsection{Baseline Subtraction Ablation}
\label{app:subtraction-ablation}
\label{app:csd_subtraction}

Baseline subtraction is required because the target prompt has its own digit distribution. In clean target runs without hidden-state transfer, the suffix strongly favors the token \texttt{1}. Across the evaluated target runs, the median ratio between the largest and smallest digit probabilities is approximately fifteen. Reading the patched logits directly would therefore combine evidence supplied by the transferred state with a fixed preference supplied by the stripped prompt.

Subtracting
\(
z_{\mathrm{base}}
\)
removes this prompt contribution in logit space before the digit and residual scores are normalized. The operation is applied independently at every vocabulary entry, so a source state must raise a candidate relative to the clean target run for that candidate to receive positive corrected evidence. This is especially important for early layers, where the transferred state may carry only weak answer information and the target-prompt preference could otherwise determine the CSD argmax.

We repeat the full diagnostic without subtraction by replacing \Cref{eq:csd_corrected_logits} with

\begin{equation}
    z^{\ell}_{\mathrm{CSD,nosub}}
    =
    z^{\ell}_{\mathrm{patch}}.
    \label{eq:csd_no_subtraction}
\end{equation}

The availability-before-readiness ordering remains present across all 16 evaluated models under this ablation. FPCL is unchanged because it is defined entirely by Probing. FJC and the outcome proportions shift because the uncorrected CSD trajectories inherit the digit preference of the target prompt. The ablation therefore preserves the central separation between Availability and Readiness while showing that subtraction is needed for calibrated comparisons among digit candidates and for stable sample-level outcome assignment.

The default analysis uses baseline-subtracted logits throughout. This choice prevents the fixed target-prompt distribution from being counted as evidence carried by the source hidden state. It also keeps CSD focused on the contribution that changes when a particular state is transferred, which is the quantity required by the operational definition of Readiness.

\subsection{CSD Validity Checks}
\label{app:csd-validity}

We conduct two supplementary experiments to assess the validity of CSD.

First, we preserve the original prompt and accumulated hidden states, then block attention from question positions to code positions after each cut layer. This evaluates downstream decoding without hidden-state transfer. The resulting layer-wise accuracy curves correlate strongly with CSD, with a mean Pearson correlation of $r=0.978$ (\Cref{fig:csd-validity}a).

Second, we keep the source hidden states fixed and test three alternative target prompts. Their layer-wise accuracy curves remain strongly correlated with those under the original target prompt, with a mean Pearson correlation of $r=0.971$ (\Cref{fig:csd-validity}b). This supports the consistency of the decoding profile across the tested prompts.

\section{Resolution Outcome Taxonomy Analysis}
\label{app:resolution}

With the implementation and reliability checks for Probing and CSD described above, we turn to the trajectory categories constructed from the paired readouts and the final model prediction. This section restates the classification rule, evaluates its sensitivity to the late CSD threshold and tail window, and reports the resulting trajectory distributions. We then examine trajectories without FJC and the separate \textsc{No Brewing} category to clarify what each label does and does not imply.

\subsection{Outcome Definitions}
\label{app:outcome-def}

Let $\hat{t}_{\mathrm{out}}$ denote the final model prediction. For a trajectory without FJC, we summarize late CSD behavior over the final quarter of the model:
\begin{equation}
\overline{\Phi}_{C}^{W}(t)
=
\frac{1}{|W|}\sum_{\ell\in W}\Phi_{C}^{\ell}(h^{\ell})[t],
\label{eq:tail-csd}
\end{equation}
where $W=\{\ell\mid \ell\geq \lfloor 3L/4\rfloor\}$. Writing $F\!=\!\operatorname{FJC}(S)$ and $\hat\Phi\!=\!\max_{t\in\mathcal{D}}\overline{\Phi}_{C}^{W}(t)$, the outcome rule is
{\small
\begin{equation}
\operatorname{Outcome}(S)=
\begin{cases}
\mathrm{Resolved},
& F\neq\varnothing
\;\land\; \hat{t}_{\mathrm{out}}=t^{*}\\
\mathrm{Overprocessed},
& F\neq\varnothing
\;\land\; \hat{t}_{\mathrm{out}}\neq t^{*}\\
\mathrm{Misresolved},
& F=\varnothing
\;\land\; \hat\Phi\geq 0.5\\
\mathrm{Unresolved},
& F=\varnothing
\;\land\; \hat\Phi< 0.5
\end{cases}
\label{eq:outcome-rule}
\end{equation}}
The four outcomes apply only when FPCL is defined. A trajectory with $\operatorname{FPCL}(S)=\varnothing$ is assigned to \textsc{No Brewing} and is reported outside the four outcome percentages. FPCL and FJC are defined by argmax correctness, and the value $0.5$ enters only when separating Misresolved from Unresolved trajectories through late CSD concentration.

Resolved and Overprocessed trajectories both reach FJC, so both contain a hidden state at which Probing and CSD are jointly correct. Their distinction concerns preservation after Readiness. A Resolved trajectory retains a correct final prediction, while an Overprocessed trajectory moves away from a jointly correct answer state before the output is produced. The classification does not require joint correctness to remain stable after its first attainment.

Misresolved and Unresolved trajectories reach FPCL but do not reach FJC within the evaluated depth. The late CSD rule separates sustained concentration on one digit from the absence of a comparably concentrated candidate. In the evaluated records, the candidate selected for Misresolved trajectories is an incorrect digit. This distinction describes how an incomplete transition develops near the output and does not treat either outcome as having reached Readiness.

\subsection{Threshold and Tail-Window Robustness}
\label{app:thresholds}
\label{app:taxonomy-robust}

The distinction between Misresolved and Unresolved depends on the late CSD concentration threshold and the selected tail window. We therefore test whether the reported outcome structure changes abruptly when either choice is varied. Resolved and Overprocessed are determined by FJC existence and the final prediction, while \textsc{No Brewing} is determined by FPCL existence. These three categories should remain fixed under changes that affect only the late CSD rule.

We recompute the labels from the stored trajectory records without additional model runs. We vary the concentration threshold from $0.40$ to $0.60$ while retaining the final quarter as the tail window. We then vary the tail starting point from $0.65L$ to $0.85L$ while retaining the threshold of $0.5$. Each setting changes one choice at a time and uses the same Probing and CSD trajectories.

\begin{table}[tp]
\centering
\small
\resizebox{\columnwidth}{!}{
\begin{tabular}{lccccc}
\toprule
Sweep & Resolved & Overprocessed & Misresolved & Unresolved & NB \\
\midrule
CSD threshold $0.40$--$0.60$
& 0.0 & 0.0 & 6.9 & 6.9 & 0.0 \\
Tail start $0.65L$--$0.85L$
& 0.0 & 0.0 & 2.5 & 2.5 & 0.0 \\
\bottomrule
\end{tabular}}
\caption{Maximum change in outcome share, measured in percentage points, over the two robustness sweeps. NB denotes \textsc{No Brewing}. The four outcome percentages exclude NB. Changes in Misresolved and Unresolved have equal magnitude because the sweeps move trajectories only across their shared boundary.}
\label{tab:outcome-sweeps}
\end{table}

\Cref{tab:outcome-sweeps} shows the expected pattern. Resolved, Overprocessed, and \textsc{No Brewing} remain unchanged across both sweeps. Adjusting the concentration threshold moves mass smoothly between Misresolved and Unresolved, with no abrupt change near $0.5$. Moving the tail starting point has a smaller effect, and the combined share of the two FJC undefined outcomes remains fixed. The exact cutoff therefore controls the boundary between two forms of incomplete resolution, while the distinction between trajectories that reach FJC and trajectories that do not is unaffected.

We additionally permute outcome labels and recompute the ground-truth-free separability scores introduced in \Cref{app:gt-free}. A global permutation returns the scores to chance. We also permute labels separately within each model and task combination, which preserves the outcome frequencies of every combination while removing the relation between a trajectory and its label. Across 100 permutations, the observed separability remains above the complete shuffled range. The labels therefore retain trajectory-level structure beyond the class frequencies induced by different models and tasks.

These checks show that the taxonomy is tied to the paired diagnostics by design, while its main structure is not determined by a single late CSD cutoff. \Cref{app:causal} provides a separate test through the different intervention responses associated with the outcome categories.

\subsection{Outcome Statistics}
\label{app:outcome-stats}

\begin{figure}[H]
\centering
\includegraphics[width=\columnwidth]{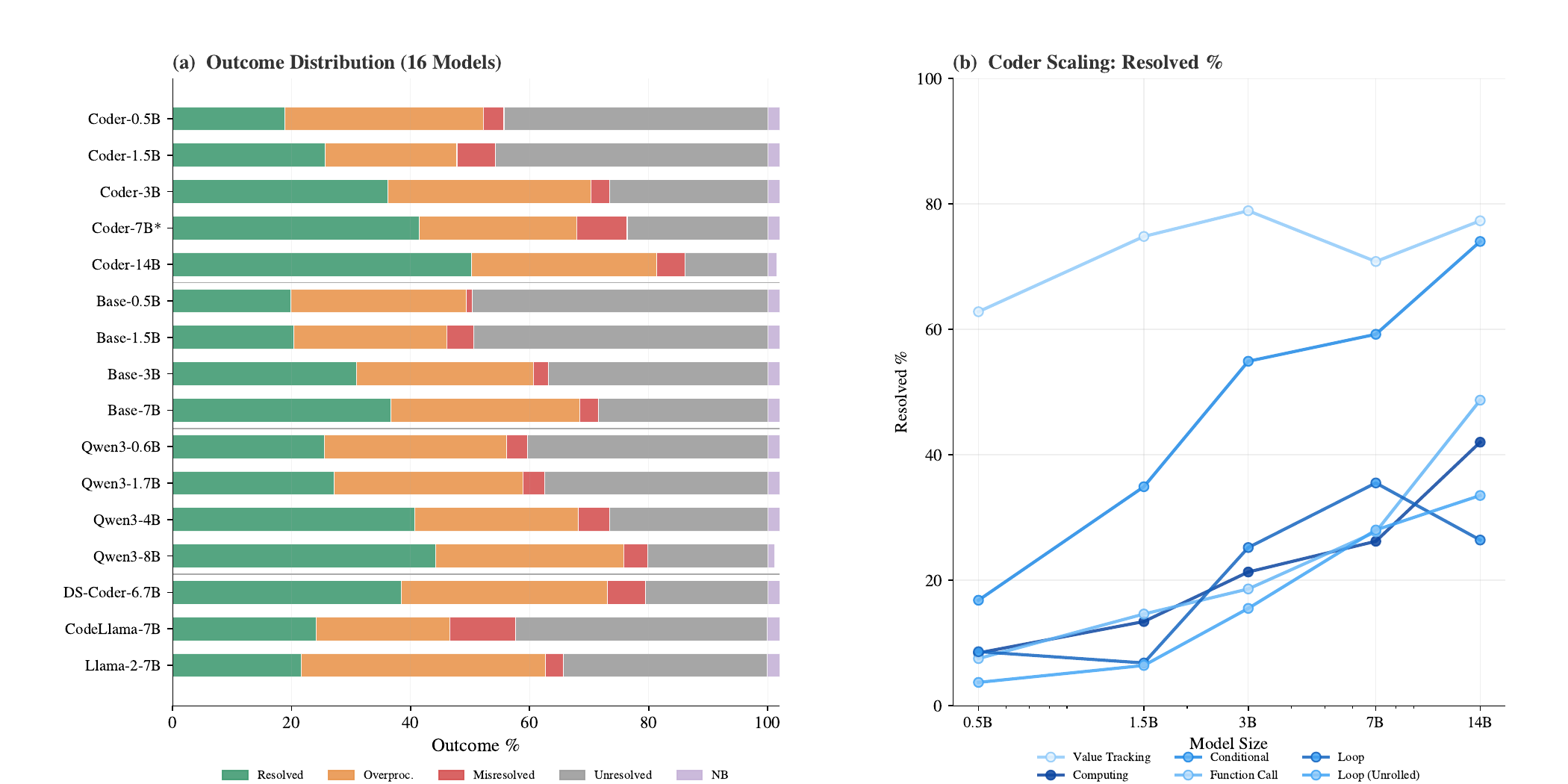}
\caption{Outcome fingerprints and difficulty responses for Qwen2.5-Coder-7B. \textbf{(a)} Flow diagram linking task families to resolution outcomes. \textbf{(b)} Resolved rate over difficulty levels.}
\label{fig:outcome-compound}
\end{figure}

The outcome labels reveal how trajectories from different code task families distribute across successful resolution, loss after Readiness, incorrect late concentration, and incomplete transition. \Cref{fig:outcome-compound} summarizes these fingerprints for Qwen2.5-Coder-7B and relates them to the primary difficulty sweep of each task family.

\begin{table}[H]
\centering
\small
\resizebox{\columnwidth}{!}{%
\begin{tabular}{lccccc}
\toprule
Task family & NB & Resolved & Overprocessed & Misresolved & Unresolved \\
\midrule
Value Tracking & 0.1 & 70.8 & 13.8 & 5.4 & 9.9 \\
Computing & 4.3 & 26.2 & 35.6 & 11.5 & 26.7 \\
Conditional & 2.5 & 59.2 & 22.7 & 10.1 & 8.0 \\
Function Call & 5.9 & 27.7 & 28.9 & 3.8 & 39.6 \\
Loop & 1.6 & 35.5 & 31.1 & 9.5 & 23.8 \\
Loop-unrolled & 1.6 & 28.0 & 26.7 & 10.4 & 34.9 \\
\bottomrule
\end{tabular}%
}
\caption{Outcome distribution (\%) for Qwen2.5-Coder-7B. NB is the \textsc{No Brewing} rate. The four outcome columns sum to $100\%$ within each task family after excluding NB.}
\label{tab:anchor-outcomes}
\end{table}

\Cref{tab:anchor-outcomes} shows that the task families differ in how their trajectories fail. Value Tracking is dominated by Resolved trajectories, with only $0.1\%$ assigned to \textsc{No Brewing}. Computing places more mass on Overprocessed than on any other outcome, indicating that many trajectories reach FJC and later lose the correct final prediction. Function Call has the largest Unresolved share, so its main loss occurs before joint correctness is reached. These comparisons use the anchor model and provide a descriptive view of how the paired diagnostic separates task-level bottlenecks.

The matched loop comparison isolates a narrower redistribution. Loop and Loop-unrolled have the same \textsc{No Brewing} rate, while Loop-unrolled has a lower Resolved rate and a higher Unresolved rate. The change in code form therefore appears primarily in whether trajectories complete the transition to Readiness within the evaluated depth. Detailed interactions with body form, iteration count, and initial offset are reported in \Cref{app:bottleneck}.

Across the 16 evaluated models, all four outcomes remain present in aggregate. Model scale and training change their relative mass, with stronger models generally showing more Resolved and fewer Unresolved trajectories, while Overprocessed remains substantial. \Cref{app:crossmodel} reports the complete scaling and model family comparisons.

\subsection{FJC-null and \textsc{No Brewing} Analysis}
\label{app:fjc-null}

\paragraph{FJC-null correctness.}
FJC is the operational landmark for Readiness under the paired diagnostic, so an undefined FJC does not require the final output to be incorrect. \Cref{tab:fjc-null-correct} reports final correctness for anchor model trajectories with defined FPCL and undefined FJC. Value Tracking reaches $41.1\%$ final accuracy in this subset, while Function Call is close to the $10\%$ single-digit baseline.
The remaining tasks lie between these cases, showing that final correctness without joint correctness is concentrated in some task families and rare in others.

\begin{table}[tp]
\centering
\small
\begin{tabular}{lrrr}
\toprule
Task family & FJC null & Correct & Accuracy \\
\midrule
Value Tracking & 620 & 255 & 41.1\% \\
Computing & 1,480 & 245 & 16.6\% \\
Conditional & 715 & 180 & 25.2\% \\
Function Call & 1,655 & 175 & 10.6\% \\
Loop & 1,330 & 270 & 20.3\% \\
Loop-unrolled & 1,805 & 270 & 15.0\% \\
\bottomrule
\end{tabular}
\caption{Final correctness among Qwen2.5-Coder-7B trajectories with defined FPCL and undefined FJC. \textsc{No Brewing} trajectories are excluded.}
\label{tab:fjc-null-correct}
\end{table}

\paragraph{Correct outputs without FJC.}
When all FJC-null trajectories are pooled, including \textsc{No Brewing}, $1{,}460$ of $8{,}255$ produce the correct final digit, which gives $17.7\%$ compared with the $10\%$ random baseline. Three analyses help characterize this excess without assigning a single explanation to every trajectory. First, correctness decreases with the primary difficulty dimension in all six task families. Function Call falls from $32.0\%$ at depth one to $4.0\%$ at depth three, while Value Tracking falls from $63.3\%$ to $27.3\%$ over the same depth range. This concentration on easier configurations is consistent with final outputs that can be supported without completing the transition measured by FJC.

Answer digit literals provide a second association. Among FJC-null trajectories, the correct group contains the answer digit more often than the incorrect group for Value Tracking and Conditional, with gaps of $27.3$ and $15.2$ percentage points. The gaps are smaller for Computing and the loop tasks, where the returned value more often depends on several updates. Literal presence therefore accounts for part of the effect in selected task families and does not explain the complete pooled excess.

The final analysis compares full-context output with the stripped CSD condition. FPCL exists for $97.3\%$ of the correct FJC-null trajectories, so linear Probing usually detects the answer at some layer. Final-layer CSD predicts the correct digit for only $28.4\%$ of the same trajectories. This gap indicates that some correct outputs depend on the original context or on a route that the stripped continuation does not reproduce. It also clarifies the scope of FJC. FJC records Readiness under the paired readout condition, while final accuracy can credit context-dependent routes that do not become jointly correct under CSD.

\paragraph{\textsc{No Brewing}.}
A \textsc{No Brewing} trajectory has undefined FPCL, which means that linear Probing does not recover the correct answer at any evaluated layer. This label does not claim that the answer is absent under every possible readout. On Qwen2.5-Coder-7B, the rate is below $2\%$ for Value Tracking and both loop forms, and it is higher for Computing at $4.3\%$ and Function Call at $5.9\%$. Within Function Call, the rate increases with call depth and reaches $9.2\%$ at depth three, linking the category to configurations where answer formation is more difficult under the linear readout.

Within the Qwen2.5-Coder scaling series, the aggregate \textsc{No Brewing} rate decreases from $10.1\%$ at 0.5B parameters to $1.4\%$ at 14B. This is a descriptive trend over five model scales. It indicates that Availability under linear Probing is detected more frequently as capability increases within this model series, while it does not identify a discrete transition at a particular scale. Together with the FJC-null analysis, this result keeps the taxonomy tied to explicit diagnostic conditions and separates failure to detect Availability from failure to reach Readiness.

\subsection{Stability Under Consecutive Correctness}
\label{app:landmark-persistence}

\begin{figure*}[t]
\centering
\includegraphics[width=\textwidth]{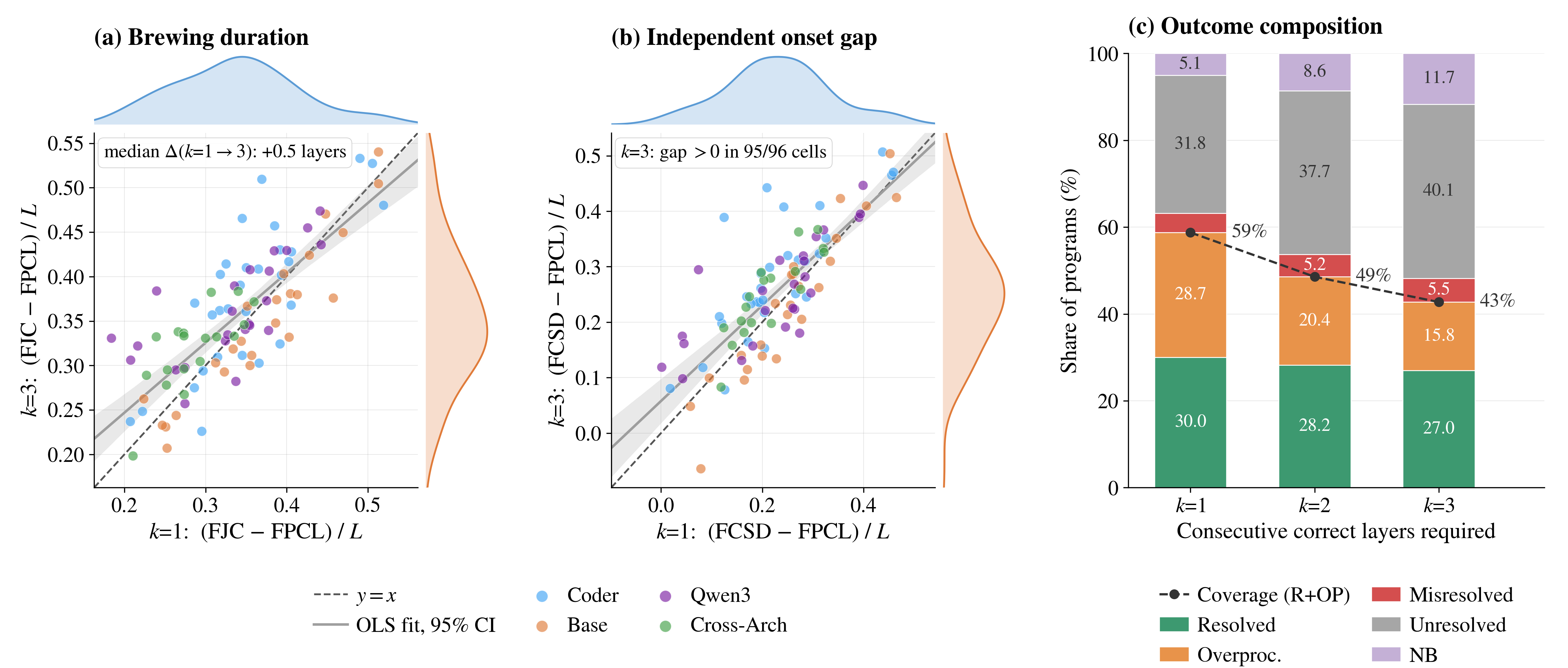}
\caption{Stability of landmarks and outcomes when correctness is required for $k$ consecutive layers. Each point is one model--task cell, colored by model group. \textbf{(a)} Normalized Brewing duration $(\mathrm{FJC}-\mathrm{FPCL})/L$ under $k=1$ and $k=3$. \textbf{(b)} Normalized gap between FPCL and the first CSD-correct layer (FCSD), measured independently of joint correctness. \textbf{(c)} Outcome composition for $k\in\{1,2,3\}$; NB denotes \textsc{No Brewing}, and coverage is the share of programs with defined FJC (Resolved plus Overprocessed). Dashed lines show $y=x$, and solid lines show OLS fits with 95\% confidence intervals.}
\label{fig:consecutive}
\end{figure*}
To assess sensitivity to transient correctness, we repeat the landmark and trajectory analyses with correctness required for two or three consecutive layers. Under these criteria, FJC remains defined for 83.9\% and 74.6\% of the trajectories with FJC under the original criterion, respectively. Brewing duration is computed only for trajectories with defined FPCL and FJC under the corresponding criterion.

The main findings remain stable under both requirements (\Cref{fig:consecutive}). Mean Brewing duration remains around 10 layers, Overprocessed trajectories remain substantial, and the difference between Loop and Loop-unrolled trajectories is preserved. These results support the stability of the aggregate findings under stricter persistence requirements, while retaining the original first-attainment definitions for the main analysis.

\section{Functional and Causal Validation}
\label{app:causal}
\label{app:functional}

The outcome analysis above separates trajectories according to FJC attainment and their subsequent behavior. These diagnostic definitions do not by themselves establish whether the corresponding answer states affect later computation in different ways. We therefore test the transition at FJC, the preservation failure in Overprocessed trajectories, the incomplete transition in Unresolved trajectories, and the late computation associated with selected failures. Unless another scope is stated explicitly, the experiments in this section use Qwen2.5-Coder-7B and held-out CUE-Bench evaluation trajectories.

\subsection{FJC Transfer Experiment}
\label{app:fjc-transfer}

FJC is the first layer at which Probing and CSD are jointly correct. Joint correctness locates the operational transition to Readiness, while the definition alone does not show that the answer state can support decoding under a separate condition. We test this by transferring states from layers around FJC into a neutral target prompt and allowing the remaining layers to generate the next token over the full vocabulary.

\begin{figure}[H]
\centering
\includegraphics[width=\columnwidth]{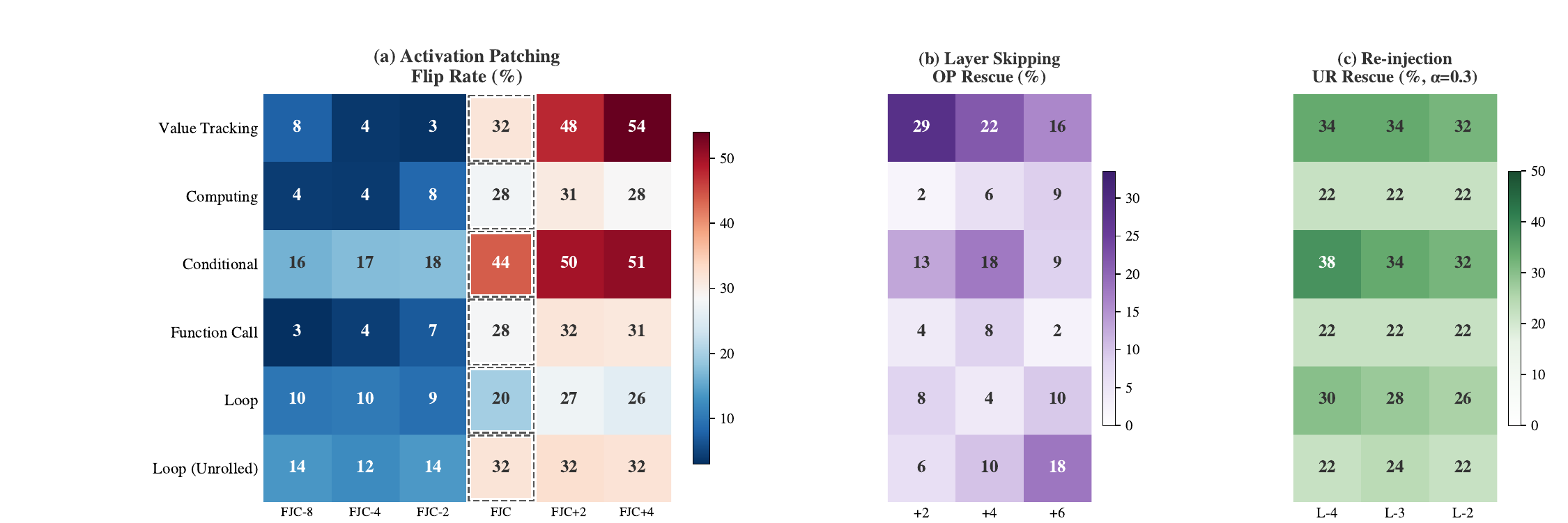}
\caption{Functional intervention results for Qwen2.5-Coder-7B. \textbf{(a)} Transfer rates around FJC. \textbf{(b)} Overprocessed recovery. \textbf{(c)} Unresolved recovery.}
\label{fig:causal-composite}
\end{figure}

For each trajectory with defined FJC, we cache the final position hidden state at offsets $\{-8,-4,-2,0,+2,+4\}$ relative to FJC. We insert each state into the final position of the neutral prompt \texttt{\# The value of x is "} at the corresponding model layer, continue through the remaining blocks, and decode with global argmax. The transfer rate is the fraction of trajectories whose generated token equals the executed answer. No restriction to the digit answer space and no CSD baseline subtraction are used in this experiment.

\begin{table}[tp]
\centering
\small
\begin{tabular}{lrrrr}
\toprule
Task family & Pre-FJC & FJC & Post-FJC & Increase \\
\midrule
Value Tracking & 5.1 & 31.5 & 50.9 & 26.5 \\
Computing & 5.6 & 27.8 & 29.6 & 22.2 \\
Conditional & 17.2 & 43.9 & 50.5 & 26.7 \\
Function Call & 4.9 & 28.3 & 31.4 & 23.4 \\
Loop & 9.9 & 19.8 & 26.6 & 9.8 \\
Loop-unrolled & 13.3 & 31.7 & 32.2 & 18.3 \\
\bottomrule
\end{tabular}
\caption{Correct-answer transfer rate in percent around FJC. Pre-FJC averages offsets $-8$, $-4$, and $-2$. Post-FJC averages offsets $+2$ and $+4$. Increase is the FJC rate minus the Pre-FJC average.}
\label{tab:fjc-transfer}
\end{table}

\Cref{tab:fjc-transfer} shows a consistent increase at FJC. Pre-FJC transfer ranges from $3$--$18\%$ at the individual offsets, while transfer at FJC ranges from $20$--$44\%$. The mean increase across task families is $21.2$ percentage points, and the rate remains elevated after FJC. This change is measured through a different target prompt and full-vocabulary decoding, so it does not follow from the FJC definition. The result supports FJC as a transition in how effectively the remaining computation can use the transferred answer state.

The size of the increase varies across task families. Loop has the smallest change, while Value Tracking and Conditional continue to improve after FJC. These differences indicate that FJC marks first attainment of Readiness and does not require the answer state to have reached its strongest later form. They also motivate testing whether a state available at FJC can counteract later processing when the final prediction becomes incorrect.

\subsection{Overprocessed Intervention}
\label{app:overprocessed-intervention}

Overprocessed trajectories reach FJC and later produce an incorrect final prediction. We test whether the FJC answer state remains useful by introducing it at a downstream layer of the same forward pass. This intervention bypasses part of the computation after FJC while retaining the original prompt and its attention context.

For a trajectory with FJC at layer $\ell$, we introduce $h^{\ell}$ at layer $\ell+k$, where $k\in\{2,4,6\}$. We compare direct replacement, norm matching, and alpha-blend injection. The alpha-blend state is
\begin{equation}
\widetilde{h}^{\ell+k}
=
(1-\alpha)h_{\mathrm{orig}}^{\ell+k}
+
\alpha h^{\ell},
\qquad \alpha=0.3.
\label{eq:alpha-blend}
\end{equation}
Direct replacement yields a mean recovery rate of $10.4\%$, while norm matching raises the mean to $31.5\%$. Alpha blend reaches $47.8\%$ at the nearest downstream target. The comparison indicates that direct replacement is strongly affected by the difference between the FJC state and the later residual state. Alpha blend retains the later state while reintroducing information carried at FJC.

\begin{table}[tp]
\centering
\small
\resizebox{\columnwidth}{!}{
\begin{tabular}{lrrrr}
\toprule
Task family & Rec.\ $+2$ & Rec.\ $+4$ & Rec.\ $+6$ & Res.\ ctrl $+2$ \\
\midrule
Value Tracking & 67.3 & 41.3 & 35.1 & 96.0 \\
Computing & 39.6 & 29.2 & 22.2 & 60.0 \\
Conditional & 52.2 & 40.0 & 20.0 & 92.0 \\
Function Call & 36.7 & 36.2 & 20.0 & 70.8 \\
Loop & 45.8 & 29.2 & 26.2 & 68.0 \\
Loop-unrolled & 44.9 & 32.7 & 34.1 & 76.0 \\
\bottomrule
\end{tabular}}
\caption{Overprocessed recovery under alpha-blend injection with $\alpha=0.3$. Recovery is the percentage of initially incorrect Overprocessed trajectories changed to the correct digit. The control column reports accuracy for Resolved trajectories under the same nearest-layer intervention. Fifty Overprocessed and twenty-five Resolved trajectories are selected per task, with effective counts reduced when an offset exceeds the layer range.}
\label{tab:overprocessed-recovery}
\end{table}

The nearest downstream target produces the strongest recovery in all six task families. Value Tracking reaches $67.3\%$, while the computation-intensive families range from $36.7\%$ to $45.8\%$. Recovery generally declines as the target moves farther from FJC, which is consistent with later processing changing the residual state over several layers. The Resolved controls remain more accurate than the recovered Overprocessed trajectories at the nearest target, which argues against alpha blend acting as a generic correction procedure.

The intervention provides functional support for the preservation distinction in the taxonomy. A usable answer state had formed by FJC in these trajectories, and introducing that state downstream can counteract part of the later computation associated with the wrong output. Recovery remains partial, so the result does not reduce every Overprocessed trajectory to one removable layer. It instead shows that failure after Readiness can be altered by restoring information from the earlier jointly correct state.

\subsection{Unresolved Re-injection}
\label{app:unresolved-reinjection}

Unresolved trajectories reach FPCL but never reach FJC. Probing therefore recovers the correct answer at some layer, while the unmodified forward pass does not convert that state into joint correctness. We test whether the FPCL state can still support the correct output when it is introduced closer to the end of the model.

For each Unresolved trajectory, we inject the FPCL state at layers $L-4$, $L-3$, and $L-2$ of the original prompt. We evaluate direct replacement, norm matching, alpha blend with $\alpha=0.3$, and alpha blend with $\alpha=0.5$. Direct replacement and norm matching substantially reduce the accuracy of the Resolved controls. Alpha blend with $\alpha=0.3$ provides the strongest balance between Unresolved recovery and control stability, and we use it for the layer comparison below.

\begin{table}[tp]
\centering
\small
\resizebox{\columnwidth}{!}{
\begin{tabular}{lrrrr}
\toprule
Task family & Rec.\ $L\!-\!4$ & Rec.\ $L\!-\!3$ & Rec.\ $L\!-\!2$ & Res.\ ctrl $L\!-\!4$ \\
\midrule
Value Tracking & 34.0 & 34.0 & 32.0 & 100.0 \\
Computing & 22.0 & 22.0 & 22.0 & 100.0 \\
Conditional & 38.0 & 34.0 & 32.0 & 100.0 \\
Function Call & 22.0 & 22.0 & 22.0 & 100.0 \\
Loop & 30.0 & 28.0 & 26.0 & 84.0 \\
Loop-unrolled & 22.0 & 24.0 & 22.0 & 96.0 \\
\bottomrule
\end{tabular}}
\caption{Unresolved recovery under FPCL state re-injection with alpha blend $\alpha=0.3$. Recovery is the percentage of initially incorrect Unresolved trajectories changed to the correct digit. The control column reports Resolved accuracy under the same intervention at $L\!-\!4$. Fifty Unresolved and twenty-five Resolved trajectories are selected per task.}
\label{tab:unresolved-recovery}
\end{table}

Recovery ranges from $22\%$ to $38\%$ at $L-4$ and changes little across the three target layers. The Resolved controls retain $84$--$100\%$ accuracy at the same target. This contrast shows that the intervention can help some trajectories that have not reached FJC without generally disrupting trajectories that have already resolved.

The recovery supports the distinction between Availability and Readiness for a subset of Unresolved trajectories. Their FPCL states contain answer information that can contribute to the correct prediction when introduced near the output, although the original computation does not complete the transition. Most Unresolved trajectories remain incorrect after re-injection. The category therefore includes cases that need transformations beyond a late copy of the FPCL state, and the result should not be read as evidence that all Unresolved trajectories require only additional depth.

\subsection{Late-Layer Sparsity Analysis}
\label{app:late-sparsity}

The interventions above establish different functional responses for Overprocessed and Unresolved trajectories. We next examine whether Overprocessed trajectories also show a descriptive difference in their late residual states. We measure Hoyer sparsity of the final position hidden state at every layer,
\begin{equation}
H(x)
=
\frac{\sqrt{d}-\lVert x\rVert_{1}/\lVert x\rVert_{2}}
{\sqrt{d}-1},
\label{eq:hoyer}
\end{equation}
where $d$ is the hidden dimension. Higher values indicate that activation magnitude is concentrated in fewer coordinates.

\begin{table}[tp]
\centering
\small
\resizebox{\columnwidth}{!}{
\begin{tabular}{lrrrrr}
\toprule
Task family & $N_{\mathrm{OP}}$ & $N_{\mathrm{Res}}$ & OP peak & Max gap & $\Delta H$ \\
\midrule
Value Tracking & 560 & 2,865 & 27 & 27 & 0.005 \\
Computing & 1,380 & 1,015 & 27 & 24 & 0.031 \\
Conditional & 895 & 2,340 & 27 & 24 & 0.018 \\
Function Call & 1,100 & 1,055 & 27 & 24 & 0.049 \\
Loop & 1,240 & 1,415 & 27 & 24 & 0.036 \\
Loop-unrolled & 1,065 & 1,115 & 27 & 22 & 0.006 \\
\bottomrule
\end{tabular}}
\caption{Hoyer sparsity comparison between Overprocessed and Resolved trajectories in Qwen2.5-Coder-7B. OP peak is the layer with the highest mean Overprocessed sparsity. Max gap is the layer with the largest Overprocessed minus Resolved difference, and $\Delta H$ is its magnitude.}
\label{tab:sparsity}
\end{table}

All six task families reach their maximum Overprocessed sparsity at the final layer. Since this final layer increase also appears in Resolved trajectories, the shared increase is not evidence for Overprocessing. The more informative comparison lies immediately before the output. Function Call has the largest Overprocessed minus Resolved difference at $0.049$, followed by Loop at $0.036$ and Computing at $0.031$. Value Tracking and Loop-unrolled show much smaller gaps.

This analysis is observational and does not identify which model component changes the answer. It shows that some task families with substantial Overprocessed mass also develop a distinct late residual pattern. We therefore treat sparsity as a descriptive correlate and use direct layer and component analysis for the stronger localization test below.

\subsection{Overprocessed Rewrite Localization}
\label{app:overprocessed-rewrite}

We localize the late change in an anchor model subset containing 112 Overprocessed trajectories. The logit lens analysis is restricted to 84 trajectories whose final wrong output is a digit, since the correct and competing answers can then be compared in the shared answer space. The remaining 28 trajectories produce non-digit outputs and are excluded from this analysis.

At every layer, we decode the residual state through the final normalization and unembedding. We also decompose each update as
\begin{equation}
h^{\ell}
=
h^{\ell-1}+a^{\ell}+m^{\ell},
\label{eq:residual-decomposition}
\end{equation}
where $a^{\ell}$ and $m^{\ell}$ are the attention and MLP contributions. The reconstruction error is below $10^{-4}$. We define the decisive crossover as the first layer after which the competing digit remains ahead of the correct digit through the output.

A decisive crossover is identified for 82 trajectories. Sixty-five of these crossovers occur in layers 22 through 27, with 25 at layer 22 and a median crossover at layer 22. At layers 22 and 23, all 32 trajectories in the local peak are assigned to attention by the contribution comparison. Across the complete subset with a decisive layer, 53 are attention attributed and 29 are MLP attributed. These counts locate where the competing digit overtakes the correct digit, while they remain coupled to the crossover rule and do not by themselves establish a functional role.

We therefore remove only the component written along the readout direction
\begin{equation}
d
=
\gamma\odot
\bigl(W_{U}[\hat{t}_{\mathrm{wrong}}]-W_{U}[t^{*}]\bigr),
\label{eq:readout-direction}
\end{equation}
where $\gamma$ is the final normalization scale and $W_{U}$ is the unembedding matrix. We apply the projection to the attention or MLP contribution at the crossover layer, and we compare it with a random direction. We also apply the attention projection cumulatively from the crossover layer through layer 27.

\begin{table}[tp]
\centering
\small
\resizebox{\columnwidth}{!}{
\begin{tabular}{lrrrrr}
\toprule
Subset & $n$ & Attn at $\ell$ & Attn $\ell$--27 & Rand at $\ell$ & MLP at $\ell$ \\
\midrule
All trajectories & 82 & 13.4 & 15.9 & 0.0 & 15.9 \\
Attn attributed & 53 & 17.0 & 18.9 & 0.0 & 9.4 \\
Layers 22--23 & 32 & 3.1 & 6.2 & 0.0 & 0.0 \\
\bottomrule
\end{tabular}}
\caption{Final output recovery in percent after removing a component along the competing minus correct readout direction. The layer $\ell$ is the decisive crossover layer for each trajectory.}
\label{tab:directional-ablation}
\end{table}

For attention attributed trajectories, removing the attention contribution recovers $17.0\%$ at the crossover layer and $18.9\%$ over the late segment. Applying the same projection to the MLP recovers $9.4\%$, while the random direction recovers none. The random control rules out recovery from an arbitrary reduction in activation magnitude. The partial effect and the small response to coarse single layer removal indicate that the competing answer is written across several late layers rather than through one isolated update.

We repeat the localization and attribution procedure across the evaluated model families. Component attribution is retained only when the last layer logit lens reproduces the model output on at least $90\%$ of the analyzed trajectories. Twelve of the sixteen models pass this gate. The late crossover appears across the model set, while the component associated with it varies among the models that support reliable attribution.

\begin{table}[tp]
\centering
\small
\begin{tabular}{lrrl}
\toprule
Model family & Attn frac. & Reliable & Description \\
\midrule
Qwen2.5 Base & 0.670 & 4/4 & Attention \\
Qwen2.5-Coder & 0.507 & 4/5 & Mixed \\
Qwen3 Base & 0.683 & 1/4 & Attention \\
Llama & 0.125 & 2/2 & MLP \\
DeepSeek-Coder & 0.250 & 1/1 & MLP \\
\bottomrule
\end{tabular}
\caption{Mean fraction of late crossovers attributed to attention within each model family. Values above $0.5$ indicate more attention attributed trajectories. Reliable models satisfy the last layer reproduction gate.}
\label{tab:component-families}
\end{table}

The anchor model result supports a late attention contribution to the competing answer for the analyzed subset. \Cref{tab:component-families} limits the scope of this interpretation. Qwen2.5 Base is attention dominated, the Qwen2.5-Coder series is mixed, and the reliable Llama and DeepSeek-Coder models are MLP dominated. Several Qwen3 models do not pass the reproduction gate, so their family estimate rests on Qwen3-8B alone. The stable conclusion is the presence of a late rewrite, while the component carrying that rewrite varies across model families and scales.

\subsection{Loop vs.\ Loop-unrolled Mechanism}
\label{app:loop-mechanism}

The paired diagnostic locates the Loop and Loop-unrolled divergence within the transition toward Readiness. We next examine the late computation associated with this difference on Qwen2.5-Coder-7B. The analysis is restricted to the dual-variable condition and uses 107 matched pairs with the same initial values, repeated updates, and final answer. This scope does not support a general component claim for other task conditions or model families.

The paired trajectories remain close through layer 21. From layer 22 onward, the Loop member increasingly favors the correct digit relative to its matched Loop-unrolled member. The Loop-minus-Loop-unrolled correct-answer logit difference reaches $2.76$ at layer 26. Decomposing the residual updates from layers 22 through 27 assigns a cumulative difference of $+1.34$ to the MLP and $-0.97$ to attention. At layer 26, the MLP difference is $+1.35$, while the attention difference is $-0.05$.

To test whether this late MLP difference contributes to the output gap, we transplant the Loop twin's MLP output into the matched Loop-unrolled run. The intervention is evaluated on 70 pairs whose Loop-unrolled member initially produces an incorrect output. Attention transfer at layer 26 and MLP transfer from an unrelated donor provide component and pairing controls.

\begin{table}[tp]
\centering
\small
\resizebox{\columnwidth}{!}{
\begin{tabular}{lrrr}
\toprule
Intervention & Logit $\Delta$ & Gap closed & Correct \\
\midrule
MLP at 25--26 from Loop twin & 0.87 & 82.6 & 18.6 \\
MLP at 26 from Loop twin & 0.47 & 59.8 & 1.4 \\
Attn at 26 from Loop twin & 0.17 & 10.9 & 0.0 \\
MLP at 26 from unrelated donor & 0.21 & 29.5 & 0.0 \\
\bottomrule
\end{tabular}}
\caption{Matched transfer from the Loop twin into an initially incorrect Loop-unrolled trajectory. Gap closed and Correct are percentages over 70 dual-variable pairs.}
\label{tab:loop-transplant}
\end{table}

Transferring the matched MLP outputs at layers 25 and 26 closes $82.6\%$ of the correct-answer logit gap and changes $18.6\%$ of the initially incorrect Loop-unrolled outputs to the correct digit. The single-layer MLP transfer closes less of the gap and changes one trajectory. Attention transfer and the unrelated MLP donor change no outputs. These controls show that the effect depends on both the component and the matched Loop trajectory.

Within Qwen2.5-Coder-7B and the dual-variable condition, the decomposition and transfer results connect the observed trajectory difference to a stronger late MLP contribution in the Loop form. The result illustrates how the paired framework can move from a difference in resolution outcomes to a selected internal computation. The cross model attribution in \Cref{app:overprocessed-rewrite} also shows why this mechanism should remain scoped to the tested setting, since related late effects can be carried by different components in other model families.

\section{Ground-Truth-Free Resolution Detection}
\label{app:gt-free}

The four resolution outcomes provide a trajectory-level description based on
FPCL, FJC, and the final model prediction. These quantities require access to
the executed answer, which is available during benchmark analysis but not
during inference-time monitoring. We therefore examine whether the internal
trajectories contain signals that distinguish resolved computations without
using the ground-truth label. This section constructs ground-truth-free
statistics from the final layers and evaluates their ability to identify
Resolved trajectories.

\subsection{Layer-wise Signals}
\label{app:gt-signals}

The proposed detection signals summarize the final portion of the Probing and
CSD trajectories, where the model approaches its final prediction. We focus on
three quantities that capture complementary aspects of the paired diagnostics.

First, we measure the uncertainty of the CSD distribution through the entropy
of the CSD probabilities over the shared target space. Lower entropy indicates
that the remaining computation concentrates on fewer candidate answers.

Second, we measure the divergence between Probing and CSD distributions using
the Jensen--Shannon divergence. A smaller divergence indicates that the two
diagnostics agree more strongly on the same hidden state.

Third, we compute the agreement rate between the Probing and CSD argmax
predictions over the selected layers. This statistic captures whether the two
readouts converge to the same candidate answer near the end of the network.

For each trajectory, all three statistics are averaged over the final 25\% of
layers. This window follows the same motivation as the late-layer CSD analysis
used for distinguishing Misresolved and Unresolved trajectories. The resulting
signals do not use the executed answer and are therefore independent of the
ground-truth label used in the outcome definition.

\subsection{Resolution Functional}
\label{app:resolution-functional}

We combine the three layer-wise signals into a single score that summarizes
late-layer resolution behavior.

Let $\hat{h}$ denote the normalized average CSD entropy, $\hat{\delta}$
denote the normalized Jensen--Shannon divergence between Probing and CSD, and
$\hat{a}$ denote the argmax agreement rate. We define the Resolution
Functional as
\begin{equation}
\rho = (1-\hat{h})(1-\hat{\delta})\hat{a}.
\label{eq:resolution-functional}
\end{equation}

A high $\rho$ value corresponds to a trajectory with concentrated CSD
predictions, strong agreement between the two diagnostics, and stable late-layer behavior. Because the score is computed only from internal trajectories,
it can be evaluated without knowing whether the final answer is correct.

The Resolution Functional is designed as an observability measure of the
internal trajectory rather than a replacement for the outcome definition.
Resolved trajectories are expected to exhibit stronger agreement and lower
uncertainty, while different failure outcomes may retain partial signals that
reflect their distinct internal behaviors.

\subsection{Detection Performance}
\label{app:gt-detection}

\begin{figure}[tp]
  \centering
  \includegraphics[width=0.85\columnwidth]{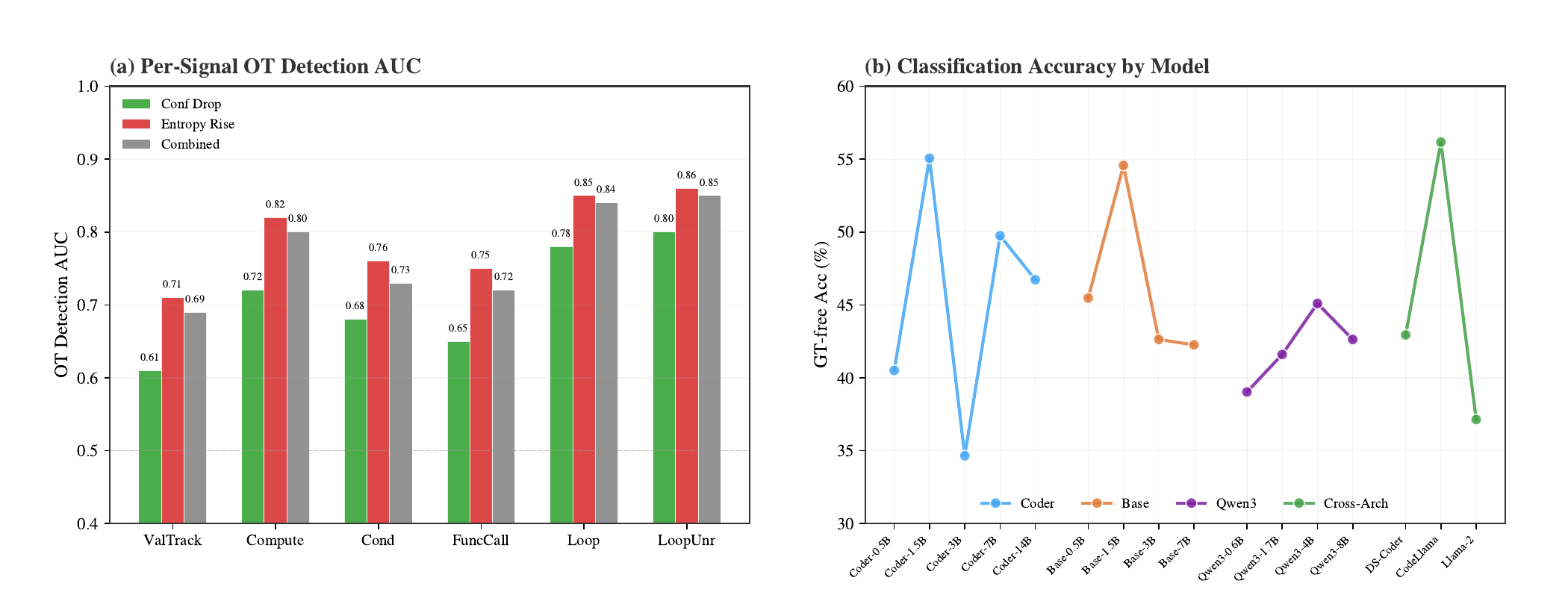}
  \caption{Ground-truth-free resolution detection. Internal trajectory signals distinguish resolution outcomes without access to ground-truth labels.}
  \label{fig:gtfree}
\end{figure}

We evaluate whether the Resolution Functional can distinguish Resolved
trajectories from the remaining outcomes without access to ground truth during
prediction. The evaluation is performed over all models and task families using
the same held-out trajectories used for the main analysis.

Across the evaluated settings, $\rho$ achieves an AUC of 0.850 for separating
Resolved trajectories from the remaining outcomes. This result shows that
late-layer internal signals contain information about whether a trajectory has
successfully completed resolution.

The detection performance varies across task families because different code
structures produce different failure distributions. Tasks with larger
fractions of incomplete transitions or later preservation failures expose
different internal patterns, which affects how easily the final-layer signals
separate outcomes.

Importantly, the detector does not use the executed answer, FPCL, or FJC when
computing $\rho$. It therefore provides a complementary view of the paired
diagnostic framework by showing that internal resolution behavior leaves
observable signatures even when the final label is unavailable.

\subsection{Lens Ablation}
\label{app:gt-ablation}

The Resolution Functional combines signals from both Probing and CSD. We
therefore examine whether both readouts contribute to the final detection
performance.

We compare the full functional score against variants that remove one of the
two diagnostic perspectives. The Probing-only variant measures late-layer
confidence from the external readout, while the CSD-only variant measures
late-layer behavior from the model's remaining computation. The full score
combines information from both conditions.

Removing either diagnostic reduces the ability to distinguish resolution
outcomes, especially when separating trajectories that have similar final
confidence but differ in whether the two readout conditions agree. The paired
version retains information about the relationship between external
recoverability and model-internal decoding.

These ablations support the role of the paired diagnostic framework. The
ground-truth-free signal does not arise from a single confidence statistic, but
from comparing how the same hidden state behaves under the two decoding
conditions.

\section{Cross-Task Bottleneck Analysis}
\label{app:bottleneck}

The outcome distributions in \Cref{app:resolution} show that different code task families produce different trajectory patterns. We further analyze how controlled difficulty changes redistribute trajectories across the four outcomes. Unless otherwise stated, all analyses in this section use Qwen2.5-Coder-7B as the anchor model. The goal is to locate whether increased difficulty mainly affects the transition toward Readiness or the preservation of a state after Readiness.

\subsection{Per-Task Outcome Distributions}
\label{app:per-task-outcomes}

\begin{figure}[tp]
\centering
\includegraphics[width=\columnwidth]{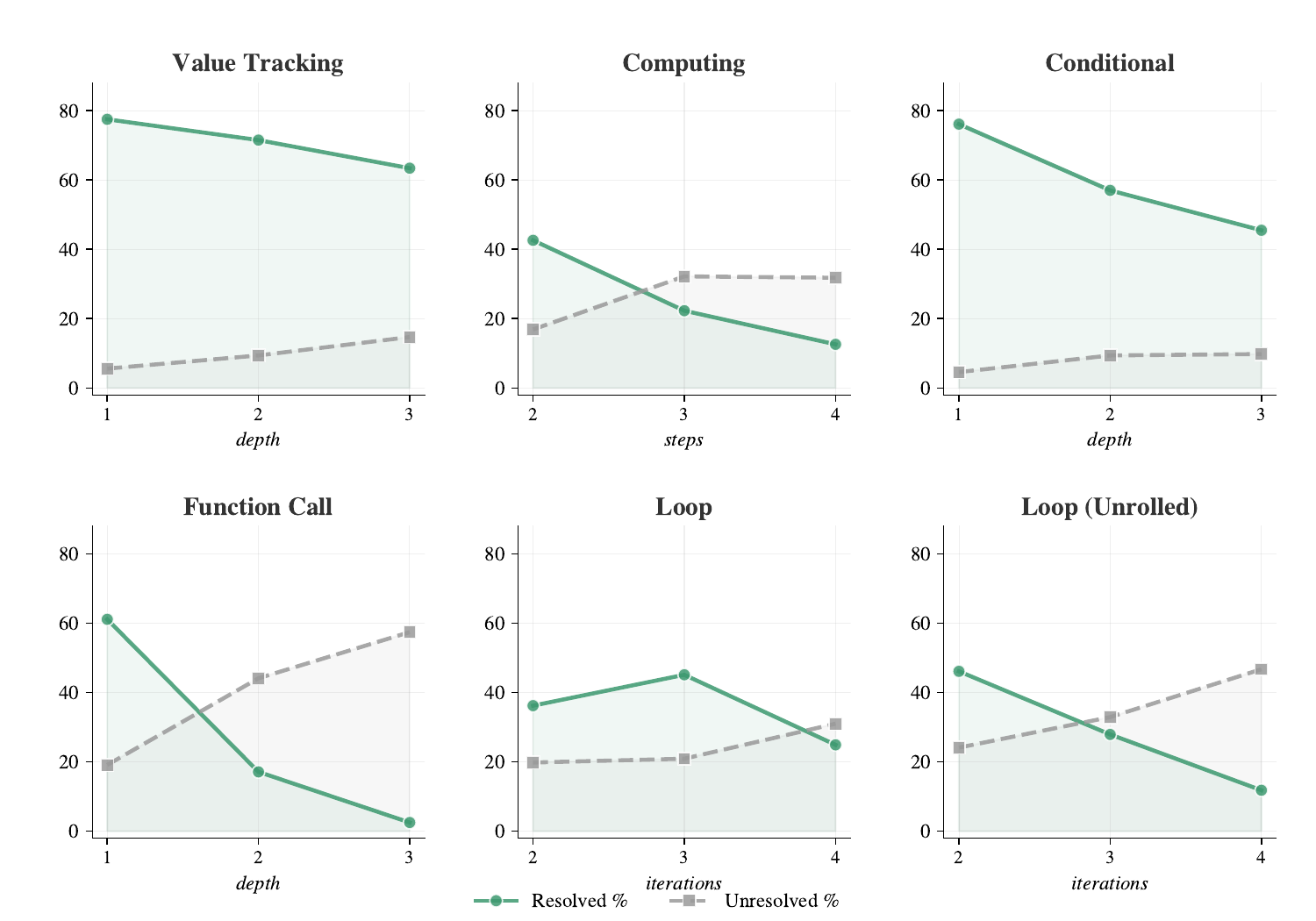}
\caption{Outcome distributions across difficulty levels for the six task families on Qwen2.5-Coder-7B. Different task families redistribute trajectories differently as difficulty increases.}
\label{fig:difficulty}
\end{figure}

\Cref{fig:difficulty} summarizes the outcome distributions across difficulty levels for the six task families. Each sweep changes one primary difficulty dimension while marginalizing over the remaining dimensions. The resulting distributions show that similar reductions in final performance can correspond to different changes in internal trajectories.

Value Tracking remains dominated by Resolved trajectories across difficulty levels. Increasing propagation depth changes the outcome distribution only moderately, suggesting that value propagation without transformation is less likely to block the transition from Availability to Readiness under the tested settings.

Computing and Function Call exhibit different failure responses despite both requiring multi step computation. Increasing arithmetic steps in Computing mainly increases Overprocessed trajectories. In contrast, increasing function depth in Function Call mainly increases Unresolved trajectories. These patterns indicate that the two task families fail at different stages after difficulty increases.

Conditional shows outcome changes associated with the form of branch reasoning. Different condition types redistribute trajectories among the failure outcomes, suggesting that the representation of a predicate can affect whether the model reaches a stable resolution.

The Loop family provides a controlled view of repeated state updates. Increasing iteration count changes the balance between successful resolution and incomplete trajectories. Detailed comparison between Loop and Loop-unrolled programs is provided in \Cref{app:loop-comparison}.

\subsection{Difficulty Sweeps}
\label{app:difficulty-sweeps}
\label{app:brewing-duration}

The difficulty sweeps isolate how individual structural dimensions affect the paired diagnostic trajectory. For each task family, we vary one dimension while keeping the remaining dimensions unchanged in distribution. This prevents changes in outcome frequency from being attributed only to a different mixture of task configurations.

\paragraph{Computing.}

For Computing, increasing the number of arithmetic steps shifts trajectories toward Overprocessed outcomes. The FPCL distribution changes only slightly across step counts, indicating that answer information becomes available at similar depths. The main change occurs after the answer reaches joint correctness, where later computation increasingly fails to preserve the correct state.

\paragraph{Function Call.}

Function Call shows a different pattern. Increasing function depth reduces the fraction of trajectories reaching FJC and increases Unresolved outcomes. The answer remains recoverable in many cases, while fewer trajectories complete the transition from Availability to Readiness. This places the dominant difficulty within the transition rather than after a ready state has been reached.

\paragraph{Loop and Loop-unrolled.}

Loop and Loop-unrolled maintain the same computational result by construction, allowing the effect of code form to be examined separately from execution content. Across difficulty levels, the two forms show similar trends in some settings while maintaining different outcome proportions. The largest difference appears in configurations requiring coordinated updates of two variables, where Loop-unrolled produces more Unresolved trajectories.

\subsection{FPCL and FJC Trends}
\label{app:landmark-trends}

We further examine FPCL and FJC under the difficulty sweeps to separate earlier answer formation from later answer use.

Across task families, difficulty does not affect FPCL and FJC in the same way. For Computing, additional arithmetic steps leave FPCL relatively stable while reducing the fraction of trajectories that preserve joint correctness. This matches the increase in Overprocessed outcomes.

For Function Call, increasing depth delays successful FJC attainment more than it changes FPCL. The model can still form a linearly recoverable answer state, while the remaining computation becomes less likely to decode that state.

These patterns show that difficulty does not correspond to a single internal failure mode. The paired diagnostics separate cases where the answer becomes available but fails to become ready from cases where a ready state is later altered.

\subsection{Loop vs.\ Loop-unrolled Comparison}
\label{app:loop-comparison}

Loop and Loop-unrolled provide the strictest matched comparison in CUE-Bench. The two forms share initial values, update operations, iteration counts, and verified answers. Their difference is limited to whether repeated updates are expressed through loop syntax or explicit sequential statements.

\begin{figure}[tp]
\centering
\includegraphics[width=\columnwidth]{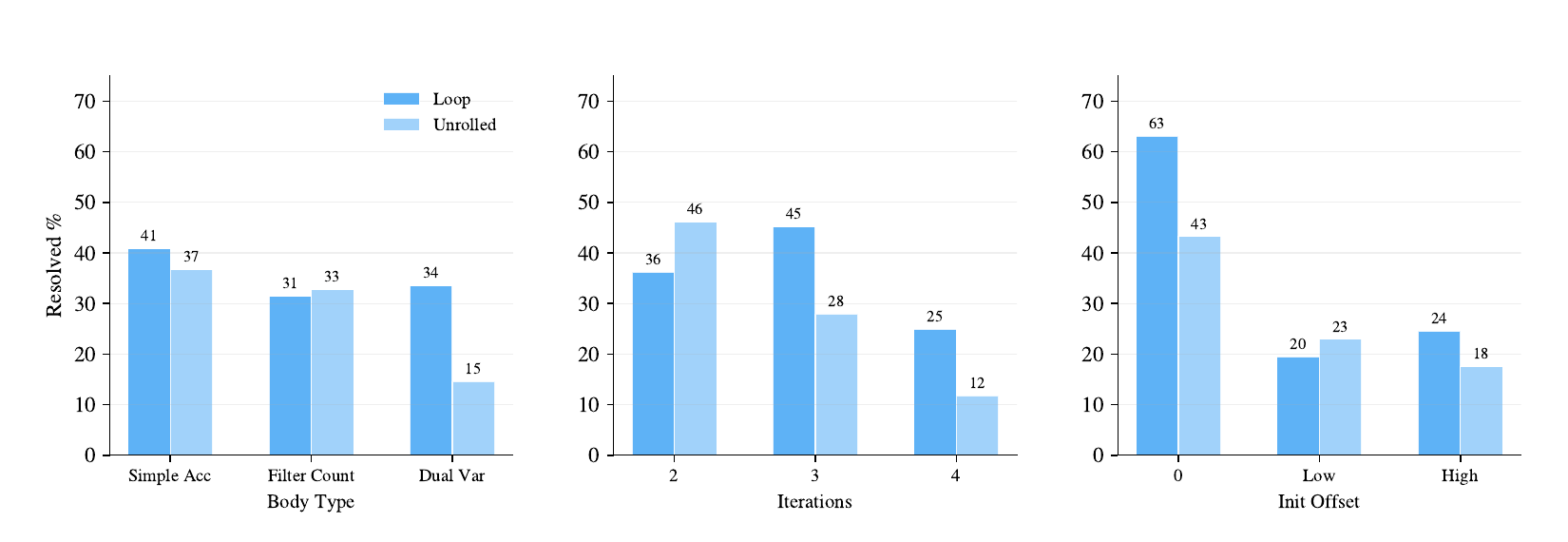}
\caption{Matched Loop and Loop-unrolled comparison on Qwen2.5-Coder-7B. The two forms reach Availability at comparable stages, while Loop-unrolled contains more trajectories that fail to complete the transition toward Readiness.}
\label{fig:loop-compare}
\end{figure}

Across the matched comparison, the difference appears after FPCL rather than at the initial formation of answer information. Loop reaches FJC more frequently, while Loop-unrolled contains more trajectories classified as Unresolved. This difference indicates that the code form changes how often an available answer state becomes usable by the model's remaining computation.

The effect is strongest in the dual-variable condition. In this setting, repeated updates require maintaining two coupled values across iterations. Loop-unrolled removes the explicit loop structure while preserving the same computation, yet the resulting trajectories more often fail before reaching Readiness.

This analysis remains limited to Qwen2.5-Coder-7B and the controlled task settings in CUE-Bench. The purpose of this comparison is to show how the paired diagnostic can localize a difference between equivalent computations rather than to assign a universal mechanism to all code forms.

\end{document}